\documentclass[11pt]{article}

\usepackage{acl}

\usepackage{times}
\usepackage{latexsym}
\usepackage{amsfonts}

\usepackage[T1]{fontenc}

\usepackage[utf8]{inputenc}

\usepackage{microtype}

\usepackage{inconsolata}

\usepackage{graphicx}

\usepackage{hyperref}
\usepackage{enumitem}
\usepackage{tabularx}
\usepackage{array}
\usepackage{pifont}
\usepackage{tikz}
\usepackage{booktabs}  
\usepackage{multirow}  
\usepackage{array}      
\usepackage{amsmath}

\usepackage{xcolor}
\usepackage{graphicx}
\usetikzlibrary{positioning}
\usepackage{framed}
\usepackage{fancyvrb}
\usepackage{caption}
\usepackage{subcaption}
\usepackage{float}
\usepackage{placeins} 
\usepackage{todonotes}

\newcommand{\posicon}{\textcolor{green!60!black}{\ding{51}}} 
\newcommand{\negicon}{\textcolor{red}{\ding{55}}}

\usetikzlibrary{shapes.geometric, arrows}
\tikzstyle{startstop} = [rectangle, rounded corners, minimum width=2.5cm, minimum height=0.8cm,text centered, draw=black, fill=white]
\tikzstyle{process} = [rectangle, minimum width=2.5cm, minimum height=0.8cm, text centered, draw=black, fill=white]
\tikzstyle{decision} = [diamond, aspect=2, text centered, draw=black, fill=white]
\tikzstyle{arrow} = [thick,->,>=stealth]

\definecolor{tfNostalgia}{RGB}{102,153,204}   
\definecolor{tfContrast}{RGB}{204,102,130}    
\definecolor{tfAnchoring}{RGB}{70,130,180}    
\definecolor{tfContinuity}{RGB}{60,170,140}   
\definecolor{tfPrimacy}{RGB}{140,110,200}     
\definecolor{tfRecency}{RGB}{200,140,90}      
\definecolor{tfUrgency}{RGB}{220,85,60}       
\definecolor{tfSkeptical}{RGB}{130,130,130}   

\newcommand{\tflabel}[2]{%
\tikz[baseline=-0.6ex]{
  \node[
    fill=#2,
    rounded corners=2pt,
    inner sep=2pt,
    font=\tiny,
    text=white
  ] {#1};
}%
}
\newcommand{\tfsep}{\hspace{2pt}}
\newcommand{\tfsentgap}{\hspace{1em}}

\definecolor{lightbrown}{RGB}{139, 69, 19}

\definecolor{darkblue}{RGB}{5, 10, 150}     
\definecolor{lightblue}{RGB}{0, 160, 245} 

\newcommand{\lightbrowntext}[1]{\textcolor{lightbrown}{#1}}
\newcommand{\darkbluetext}[1]{\textcolor{darkblue}{#1}}
\newcommand{\lightbluetext}[1]{\textcolor{lightblue}{#1}}
\definecolor{contextgray}{RGB}{90,90,120}
\newcommand{\contexttext}[1]{\textcolor{contextgray}{#1}}

\title{Uncovering Temporal Framing in the News 
}

\author{
\textbf{Tarek Mahmoud}$^1$,
\textbf{Veronika Solopova}$^{2,3}$,
\textbf{Premtim Sahitaj}$^{2,3}$,
\textbf{Ariana Sahitaj}$^{2,3}$, \\
\textbf{Max Upravitelev}$^{2,3}$,
\textbf{Mervat Abassy}$^1$,
\textbf{Hana Fatima Shaikh}$^4$,
\textbf{Neda Foroutan}$^2$, \\
\textbf{Vera Schmitt}$^{2,3}$,
\textbf{Preslav Nakov}$^1$ \\
$^1$MBZUAI, UAE \quad
$^2$Technische Universität Berlin, Germany \\
$^3$German Research Center for Artificial Intelligence (DFKI), Germany \\
$^4$University of Maryland, USA \\
\texttt{\{tarek.mahmoud, preslav.nakov\}@mbzuai.ac.ae}
}

\begin{document}
\maketitle
\begin{abstract}
Temporal language does more than place events on a timeline. In news discourse, references to the past, present, and future can function as rhetorical devices that shape interpretation and persuasion. Here, we study \emph{temporal framing}, defined as the persuasive use of time-related language to structure meaning rather than to report chronology. We propose a taxonomy of eight temporal frames grounded in prior work on temporality and framing, and we realize it through expert annotation of a multilingual news corpus. The resulting dataset includes 458 English and German news articles, with over 2K temporally framed sentences and approximately 3K temporal framing annotations identified from a corpus of more than 20K sentences. We analyze frame prevalence, co-occurrence patterns, and lexical cues, and evaluate temporal framing detection using supervised fine-tuning and zero-shot classification. Our experiments show that temporal framing is learnable at the sentence level, with supervised models substantially outperforming zero-shot approaches. We publicly release the corpus to support future research on temporal framing: 
\url{https://mbzuai-nlp.github.io/temporal-framing/}.

\end{abstract}

\begin{figure*}[!t]
\centering

\begin{subfigure}[t]{\textwidth}
\centering

\fbox{%
\parbox{0.97\textwidth}{%
\small
\setlength{\parindent}{0pt}
\setlength{\parskip}{0.5em}

{\centering \textbf{Violence Has Been Normalized}\par}

During the misnamed and mostly preposterous debate between Kamala Harris and Donald Trump, a moderator fact-checked Trump's claim that crime is up.

In contrast to Trump's claim, moderator David Muir said that the FBI reports that crime is down, a claim that likely struck every viewer as obviously wrong.

Shoplifting was not a way of life before lockdowns.\tfsep
\tflabel{Nostalgia}{tfNostalgia}\tfsep
\tflabel{Temporal Anchoring}{tfAnchoring}\tfsep
\tflabel{Temporal Contrast}{tfContrast}\tfsentgap
Most cities were not demographic minefields of danger around every corner.\tfsep
\tflabel{Nostalgia}{tfNostalgia}\tfsep
\tflabel{Temporal Contrast}{tfContrast}\tfsentgap
There was no such thing as a drugstore with nearly all products behind locked Plexiglas.\tfsep
\tflabel{Nostalgia}{tfNostalgia}\tfsep
\tflabel{Temporal Contrast}{tfContrast}

We weren't warned of spots in cities, even medium-sized ones, where carjacking was a real risk.\tfsep
\tflabel{Nostalgia}{tfNostalgia}\tfsep
\tflabel{Temporal Contrast}{tfContrast}

It is wildly obvious that high crime in the U.S. is endemic, with ever less respect for person and property.\tfsep
\tflabel{Continuity}{tfContinuity}\tfsep
\tflabel{Primacy}{tfPrimacy}\tfsentgap
As for the FBI's statistics, they're worth about as much as most data coming from federal agencies these days.\tfsep
\tflabel{Temporal Contrast}{tfContrast}

}%
}

\caption{\textbf{Annotated excerpt illustrating temporal framing.} 
The complete example appears in Appendix~\ref{sec:full_annotated_example}.}
\label{fig:temporal_framing_inline_example}

\end{subfigure}

\vspace{0.8em}

\begin{subfigure}[t]{\textwidth}
\centering

\small
\setlength{\tabcolsep}{6pt}
\renewcommand{\arraystretch}{1.15}

\begin{tabularx}{\textwidth}{>{\bfseries}l X X}
\toprule
Temporal Frame & \textbf{Short Definition} & \textbf{Example} \\
\midrule
Primacy &
\emph{Significance attributed to being first in time.} &
``The first to discover the cure will lead the world.'' \\
\midrule
Recency &
\emph{Significance attributed to the most recent events.} &
``Today's figures matter most for understanding public opinion.'' \\
\midrule
Urgency &
\emph{Emphasis on limited time or imminent consequences or threats.} &
``Scientists warn the next 48 hours are critical.'' \\
\midrule
Temporal Anchoring &
\emph{Framing the discussion through the lens of past events.} &
``We live in a post-9/11 world where security comes first.'' \\
\midrule
Nostalgia &
\emph{Invocation of a cherished past.} &
``We must act now to bring back the good old days.'' \\
\midrule
Temporal Contrast &
\emph{Juxtaposition of ``then'' versus ``now.''} &
``Once a booming hub, now a city in decline.'' \\
\midrule
Continuity &
\emph{Persistence across time.} &
``The economy has been climbing steadily for a decade thanks to policy changes.'' \\
\midrule
Skeptical &
\emph{Casting doubt about the future.} &
``The government's plan may collapse under financial pressure.'' \\
\bottomrule
\end{tabularx}

\caption{\textbf{Brief overview of the temporal framing taxonomy with definitions and examples.} For detailed definitions, additional examples, and rhetorical functions, see Appendix~\ref{sec:detailed_taxonomy}.}
\label{tab:temporal-framing-definitions}
\end{subfigure}

\caption{\textbf{An annotated excerpt and an overview of the temporal framing taxonomy.}
}
\label{fig:taxonomy_and_example}
\end{figure*}

\section{Introduction} 
\label{sec:intro}

All languages provide systematic means to express temporal relations, such as ordering, simultaneity, duration, and change over time,
whether through tense, aspect, lexical items, or discourse-level mechanisms \cite{comrie1985tense, evans2009myth}. Beyond merely locating events on a timeline, temporal expressions influence how speakers establish causality and how they invite audiences to reason about the past, the present, and the future. Linguistic choices related to time play a central role not only in describing reality, but also in framing it \cite{33702df9-9bc1-35d6-b14b-016eee94408a}.

In computational linguistics, however, time has traditionally been treated in a narrow instrumental manner, with focus on  predicting verb tense or aspect, while recent work in NLP has concentrated on temporal expression extraction and temporal reasoning \cite{tan-etal-2023-towards,10.1145/3701716.3717744}. 
Although these efforts have enabled significant progress in temporal reasoning, they typically conceptualize time as an objective property of events, and not a rhetorical resource. As a result, current NLP systems largely overlook the richer semantic and pragmatic functions that temporal language fulfills in discourse. While prior work has explored framing in NLP more broadly \cite{otmakhova-etal-2024-media,mahmoud-etal-2025-entity}, the intersection of time and framing remains underexplored. This gap limits our ability to analyze persuasive rhetorical structures along the temporal dimension, particularly in domains such as political communication, where temporal justifications and appeals are known to be especially influential \cite{solopova2023prokremlin}. Concretely, temporal framing enables fine-grained analysis of how news and political discourse construct narratives through references to the past, present, and future, supporting downstream applications such as media profiling, stance and narrative analysis, and cross-lingual comparison of framing strategies across outlets and events.

In social sciences, time is widely recognized as a powerful framing device \cite{trope2010construal}. Temporal framing categories enable systematic study of phenomena such as appeals to nostalgia, constructions of urgency, or the use of historical anchoring to legitimize policy decisions across events, outlets, and time, in diverse communicative and sociopolitical contexts globally and historically. In political discourse studies, temporal references are explicitly analyzed as rhetorical strategies that frame interpretation and legitimize authority \cite{wodak2015politics}. 

Temporal references can create urgency by compressing perceived time, convey stability by emphasizing continuity, evoke nostalgia of an idealized past, or anchor present arguments to emotionally charged historical moments. These patterns are routinely studied in domains such as political communication, crisis reporting, and health messaging, where temporal framing shapes interpretation and decision-making.
The persuasive force of temporal references lies in their relational positioning within the discourse. Expressions such as \emph{now} or \emph{for decades} gain rhetorical meaning from implied alternative timeframes. 
Temporal framing operates at a rhetorical level, emerging from how temporal references are positioned within a sentence and the broader context. This makes temporal framing challenging to capture with purely extraction-based approaches. 




Motivated by this gap, we present a computational study of temporal framing in news discourse (see Figure~\ref{fig:temporal_framing_inline_example} for an example). Our contributions are fourfold:
\begin{enumerate}
    \item We cast temporal framing as a sentence-level, multi-label task capturing the rhetorical use of temporal language beyond surface expressions.
    \item We propose a taxonomy of eight temporal frames grounded in social science research.
    \item We introduce an annotation scheme and release a multilingual news dataset of 458 articles annotated at the sentence level.
    \item We provide computational baselines for detecting temporal framing using diverse language models in zero-shot and fine-tuned settings.
\end{enumerate}

\section{Related Work}
\label{sec:related_work}



\paragraph{Framing in NLP} 

Framing has been widely studied in NLP. Early work has primarily focused on \emph{document-level framing}, where each article is assigned one or more dominant frames~\cite{card-etal-2015-media, liu-etal-2019-detecting, delBario2023}. This line of work has enabled large-scale analysis of framing trends across outlets and over time spans, supporting comparative studies of media bias and agenda setting, but abstracts away from how framing is realized locally in language and interaction. More recent work has explored \emph{entity-level}, \emph{event-level}, and \emph{narrative} framing, examining how framing emerges through the roles, attributes, and relationships assigned to entities and events~\cite{stammbach-etal-2022-heroes, otmakhova-etal-2024-media, Gehring2023, das-etal-2024-media, mahmoud-etal-2025-entity}. These approaches capture finer-grained framing dynamics and provide insight into how meaning is constructed within texts, including the interplay between actors and events in discourse. While these approaches capture finer-grained framing dynamics, they do not explicitly model the rhetorical use of temporal language as a framing dimension.
    
\paragraph{Temporal Analysis in NLP} 

A large body of research addresses temporal understanding through tasks such as temporal expression extraction, event ordering, timeline construction, and temporal reasoning~\cite{chu-etal-2024-timebench, islakoglu-kalo-2025-chronosense}. Benchmarks and models in this area aim to recover factual temporal structure, answering questions about when events occurred and how they relate chronologically. Although these methods advance computational representations of time, they largely treat temporal information as objective and descriptive rather than persuasive.

Recent work has also examined temporal structure in news at the corpus level. Although \citet{Cobo2025} use the term \emph{temporal framing}, their work defines temporality as the evolution of frames across time rather than as a rhetorical device expressed in language. Likewise, \citet{Card2022} analyze how news framing varies over time. Similarly, \citet{lamorte-etal-2024-timeframe} introduce an interactive system for event-centered exploration of news archives, enabling users to browse events and frames along a timeline. While these approaches capture how frames and events evolve over time, they treat time as an organizing or analytical dimension rather than a rhetorical one.


\paragraph{Temporal Framing in Social Sciences} 

Temporal framing has been extensively studied in psychology, communication, and political science. Foundational work in construal level theory shows that temporal distance systematically shapes cognition, with proximal events construed more concretely and distal events more abstractly, influencing evaluation, emotion, and decision-making~\cite{trope2010construal}. In applied domains such as health communication, temporal framing of risks and outcomes affects perceived severity and behavioral intentions~\cite{Chandran2004, kim2018time, orbell2008temporal}. In political discourse, temporal references to the past or future are used to structure interpretation, justify policy choices, and legitimize action~\cite{wodak2015politics, wagoner2025thinking}. Despite this extensive work, these insights have not yet been systematically analyzed or modeled in NLP.


Although \emph{temporal framing} is well studied in social sciences, NLP has largely overlooked time as a rhetorical device. Existing datasets and models fail to capture how authors use time beyond factual chronology. We address this gap by introducing a sentence-level temporal framing taxonomy and annotated dataset for news discourse.

\section{Temporal Framing}
\label{sec:temporal_framing}




Temporal framing refers to the rhetorical use of time-related elements to structure meaning and persuade audiences. Rather than merely situating events chronologically, it highlights, contrasts, or attributes significance to moments, durations, and trajectories to shape how issues are perceived.

To make these distinctions more precise, we introduce a taxonomy of temporal frames summarized in Figure~\ref{tab:temporal-framing-definitions} and described in Appendix~\ref{sec:detailed_taxonomy}. The taxonomy identifies eight ways temporal reference is used rhetorically, ranging from appeals to primacy, recency, or urgency to frames invoking nostalgia, continuity, or skepticism about the future. Temporal references may be accompanied by markers such as \textit{first, now, still, since, before, after, deadline, turning point, for years, pre-war, last month, next year}. The taxonomy is vital for understanding nuanced uses of time, enabling granular rhetorical analysis, and is grounded in social science research (Appendix~\ref{sec:social_science_roots}). It reflects theoretically grounded and empirically stable patterns observed during annotation rather than an exhaustive set of temporal expressions.


Temporal framing can be formalized as follows.
Let $\mathcal{F}$ denote the set of temporal frames defined by the taxonomy.
Let $D = (s_1, \dots, s_n)$ be a document consisting of $n$ sentences.
The goal of temporal framing is to learn a function
\[
f : (D, s_i) \rightarrow \mathcal{P}(\mathcal{F}),
\]
where $s_i$ is a sentence in the document and $\mathcal{P}(\mathcal{F})$ denotes the power set of temporal frames.
The output $f(D, s_i)$ corresponds to the set of temporal frames whose persuasive force is expressed through time-related language in sentence $s_i$, possibly conditioned on its surrounding context in $D$.


Identifying temporal framing requires resolving two recurring challenges: (1) distinguishing \textit{factual} chronological reports from \textit{opinionated} temporal framings, and (2) focusing on the \textit{temporal} rather than the \textit{general} sense of a frame.

\paragraph{Fact versus opinionated use.}
A central distinction in our analysis is between factual chronological reporting and opinionated temporal framing. Factual reports are verifiable statements that merely \emph{describe} events or states, such as dated statistics or chronological markers, without seeking to persuade. Opinionated framings, by contrast, are rhetorical by nature: they \emph{utilize temporal elements} to persuade readers. The same sentence can shift from a neutral description to a persuasive framing depending on whether the temporal element is presented as meaningful rather than incidental. For example, ``Inflation rose by 2\% in 2024'' reports a temporal fact, whereas ``Inflation has been rising steadily for years, signaling policy failure'' uses temporal reference to support a persuasive claim.

\paragraph{Framing in the general versus temporal sense.}
Some framings may look temporal, but they may not be such. Take \textit{primacy}: in the general sense it means superiority of quality (``\emph{This is the best policy}''), while in the temporal sense, primacy refers to being first in time (``\emph{This was the first policy of its kind}''). Similarly, “\emph{Once a booming hub, now a city in decline}” expresses temporal contrast, whereas “\emph{This economy is worse than others}” conveys a general comparison without temporal framing. Distinguishing these senses is essential to avoid mistaking non-temporal framings for temporal ones.

We provide additional examples for all frames with contrasting cases in Appendix~\ref{sec:detailed_taxonomy}.

\section{Corpus Description}
\label{sec:dataset}
\subsection{Article Selection}
\label{sec:article_selection}
We construct our corpus through a multi-stage pipeline, beginning with large-scale URL retrieval from GDELT, followed by successive article extraction, filtering, preprocessing, and sampling steps.

\paragraph{Themes}
We source article URLs from the GDELT Global Knowledge Graph. We filter records by theme to construct a diverse news corpus spanning political, economic, cultural, commercial, and crisis-related reporting published between September~2024 and September~2025.

\paragraph{Corpus Retrieval}
We restrict retrieval to a set of widely circulated English and German news outlets, yielding $\sim$2M candidate articles. From this pool, we sample 6{,}000 article URLs using stratification within each $\langle$language, outlet, month$\rangle$ stratum to avoid source imbalance and short-term news bursts.

\paragraph{Text Extraction}
From the 6{,}000 URLs, we extract full article text using Trafilatura, which captures both the article content and associated metadata. We discard failed or garbled extractions and retain only articles containing at least 200 words. We further remove exact duplicates using URL matching and normalized-text hashing, followed by near-duplicate filtering via locality-sensitive hashing.


\paragraph{Article Tagging and Sampling}
An initial annotation pass indicated that opinionated content constituted a minority of the raw news corpus. To increase coverage of persuasive temporal framing, we use LLM-assisted document-level tagging to identify likely opinion articles using GPT-4o.
We combine LLM-guided upsampling with stratified uniform sampling where we target an overall opinionated document proportion of 70\% in the final dataset. We retain a substantial portion of non-opinionated articles to account for potential noise in the LLM-generated tags. It is important to highlight that these LLM tags guided sampling only at the document-level; all sentence-level temporal framing labels were assigned by human annotators.
We report the final outlet-level sampling distribution in Table~\ref{tab:outlet_full}. We further analyze the prevalence of sentence-level human annotations in articles tagged as opinionated versus non-opinionated by the LLM, as described in Appendix~\ref{sec:llm_tagging_reliability}.

\paragraph{Sentence Segmentation}
Finally, we segment the articles into sentences with Stanza using language-specific models. We prepend titles as the first sentence if they were not already part of the body. After segmentation, we discard fragments consisting only of punctuation as well as fragments containing a single character.

\begin{table}[t]
\centering
\scriptsize
\setlength{\tabcolsep}{3pt}

\begin{tabular}{llrrr}
\toprule
Lang & Media Outlet & $n$ & $n_{\text{opinion}}$ & $n_{\text{sampled}}$ \\
\midrule
\multirow{20}{*}{English}
& Forbes & 397 & 271 & 53 \\
& Breitbart & 224 & 80 & 29 \\
& The Guardian & 385 & 119 & 28 \\
& Daily Mail & 882 & 108 & 27 \\
& CNN & 356 & 33 & 18 \\
& Zero Hedge & 104 & 70 & 17 \\
& BBC & 404 & 17 & 11 \\
& Fox News & 243 & 27 & 10 \\
& China Daily (EN) & 148 & 26 & 9 \\
& CBS News & 100 & 7 & 7 \\
& SAMAA TV & 57 & 5 & 6 \\
& Daily Pakistan (EN) & 42 & 5 & 6 \\
& NBC News & 109 & 6 & 5 \\
& Time & 28 & 10 & 3 \\
& The Epoch Times & 175 & 25 & 2 \\
& Global Times (EN) & 32 & 4 & 2 \\
& Liberty Times Net (EN) & 17 & 0 & 2 \\
& Associated Press & 10 & 2 & 1 \\
& USA Today & 2 & 1 & 1 \\
& Kronen Zeitung (EN) & 1 & 0 & 1 \\
\midrule
\multirow{15}{*}{German}
& Focus Online & 107 & 31 & 31 \\
& t-online & 242 & 18 & 28 \\
& Die Welt & 359 & 12 & 26 \\
& n-tv & 276 & 12 & 24 \\
& Süddeutsche Zeitung & 174 & 35 & 22 \\
& Tages-Anzeiger & 78 & 19 & 20 \\
& Stern & 87 & 12 & 15 \\
& Bild & 125 & 13 & 14 \\
& Der Tagesspiegel & 109 & 15 & 13 \\
& NachDenkSeiten & 13 & 13 & 11 \\
& Kronen Zeitung & 44 & 4 & 5 \\
& Epoch Times (DE) & 59 & 1 & 4 \\
& Frankfurter Allgemeine Zeitung & 70 & 28 & 3 \\
& Junge Freiheit & 5 & 4 & 3 \\
& FAZ Kaufkompass & 4 & 4 & 1 \\
\bottomrule
\end{tabular}
\caption{\textbf{Media outlet-level coverage and final sampling counts.}
$n$ = successfully extracted articles.
$n_{\text{opinion}}$ denotes LLM-tagged indicators of opinionated content, used for sampling only.
$n_{\text{sampled}}$ denotes articles retained after stratified uniform sampling over time, with upsampling of opinionated articles.}
\label{tab:outlet_full}
\end{table}


\begin{table}[t]
\centering
\small
\setlength{\tabcolsep}{4pt}
\renewcommand{\arraystretch}{1.15}
\begin{tabular}{lrrrrrr}
\toprule
Lang. & $N$ & $\alpha_{\text{det}}$ & $\overline{\alpha}_{\text{frame}}$ & min & max & Set-F1$_{+}$ \\
\midrule
EN & 1{,}934 & 0.605 & 0.494 & 0.416 & 0.590 & 0.402 \\
DE & 2{,}317 & 0.720 & 0.588 & 0.302 & 0.768 & 0.443 \\
\bottomrule
\end{tabular}
\caption{\textbf{Sentence-level inter-annotator agreement.} $\alpha_{\text{det}}$ denotes Krippendorff's $\alpha$ (nominal) for binary detection of any temporal framing. $\overline{\alpha}_{\text{frame}}$ is the mean per-frame $\alpha$ across the eight temporal frames. Set-F1$_{+}$ is computed on the positive-only subset and averaged over the sentences.}
\label{tab:iaa_summary}
\end{table}

\begin{table*}[ht]
\centering
\small
\begin{tabular}{l|rrrrrrr|rrrr}
\toprule
Lang. & Docs & Sents & Words & Chars & AVG$_s$ & AVG$_w$ & AVG$_c$ & SentAnn & Ann & AVG$_{s}$ & AVG$_{a}$ \\
\midrule
EN & 238 & 8{,}750  & 161{,}311 & 992{,}556  & 36.76 & 677.78 & 4{,}170.40 & 1{,}748 & 2{,}334 & 7.34 & 1.34 \\
DE & 220 & 11{,}909 & 163{,}381 & 1{,}147{,}357 & 54.13 & 742.64 & 5{,}215.26 &   617 &   696 & 2.80 & 1.13 \\
\midrule
\textbf{Total} 
   & 458 & 20{,}659 & 324{,}692 & 2{,}139{,}913 & 45.11 & 708.93 & 4{,}672.30 & 2{,}365 & 3{,}030 & 5.16 & 1.28 \\
\bottomrule
\end{tabular}
\caption{\textbf{Corpus statistics.} The table shows the total number of documents (Docs), sentences (Sents), words (Words), and characters (Chars) by language. 
Averages per document are provided for sentences (AVG$_s$), words (AVG$_w$), and characters (AVG$_c$). 
The table also reports the number of sentences containing at least one annotation (SentAnn), the total number of annotations (Ann), 
the average number of annotated sentences per document (AVG$_{s}$), and the average number of annotations per annotated sentence (AVG$_{a}$).}

\label{tab:final_sample}
\end{table*}

\subsection{Annotation Process}


We design an iterative annotation workflow with clear guidelines and weekly calibration sessions to ensure consistency across annotators and over time. We use the \textsc{INCEpTION} annotation platform \cite{klie-etal-2018-inception}, which supports span-based and multi-label annotation as well as adjudication and collaborative review. Appendix~\ref{sec:inception} shows an example of the annotation interface. Through repeated discussion of disagreements and refinement of guidelines, this process ensured convergence on consistent interpretations of temporal framing, improved annotator alignment, and reduced ambiguity in challenging boundary cases.


A coordinator oversaw curation and consistency for each language. The team comprised three English and four German annotators. Annotators met weekly to review disagreements and revised the taxonomy and guidelines. The initial taxonomy included two frames, \emph{momentum} and \emph{cyclicity}, later merged into \emph{continuity} after annotators noted it subsumes both. We updated guidelines to exclude quoted and indirect speech, as these cases caused disagreement over whether the persuasive voice belonged to the author or quoted source. We restricted cross-sentence framing to the anchor sentence, particularly for \emph{temporal contrast}, since annotators observed that persuasive force arises from the anchor sentence. Finally, we broadened \emph{temporal anchoring} from references to historical landmarks to rhetorical anchoring to past events, as defining such landmarks proved subjective and introduced bias, whereas rhetorical past references offered a more consistent criterion.

\paragraph{Inter-Annotator Agreement.}

We randomly assign $\sim$50 articles per language to 2--4 annotators. At the sentence level (Table~\ref{tab:iaa_summary}), Krippendorff's $\alpha$ (nominal) for binary detection of temporal framing (presence vs.\ absence) reaches 0.605 for English and 0.720 for German. These agreement levels are consistent with prior work on fine-grained framing and rhetorical annotation, where moderate agreement is typical due to the interpretive and multi-label nature of the task \cite{card-etal-2015-media,piskorski-etal-2023-semeval}.
We also compute per-frame agreement by treating each of the eight temporal frames as a binary label, yielding mean $\alpha$ values of 0.494 (EN) and 0.588 (DE). To account for the multi-label nature of frame assignment, we further report positive-only set agreement, with Set-F1$_{+}$ scores of 0.402 (EN) and 0.443 (DE). 

Appendix~\ref{sec:iaa_sent_level_framewise} reports per-frame sentence-level agreement scores, and Appendix~\ref{sec:iaa_doc_level} presents \emph{document-level} agreement analysis.

\begin{figure}[t]
    \centering
    \includegraphics[width=1\linewidth]{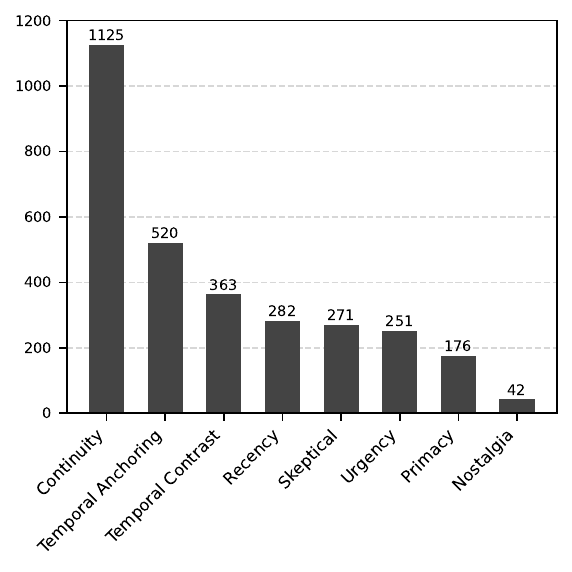}
    \caption{\textbf{Distribution of frames in the corpus.}}
    \label{fig:frames_histogram}
\end{figure}

\subsection{Corpus Analysis}
\subsubsection{Statistics}
\label{sec:statistics}
Table~\ref{tab:final_sample} presents the overall corpus statistics with a breakdown by language. While English and German documents are comparable in size, annotations are more concentrated in English, with a higher proportion of sentences labeled as temporally framed. Moreover, temporally framed English sentences contain slightly more temporal frames on average than German. Figure~\ref{fig:frames_histogram} shows the distribution of temporal frames across the dataset. \emph{Continuity}, \emph{Temporal Anchoring}, and \emph{Temporal Contrast} are the most frequent frames, indicating a dominant rhetorical focus on linking past and present or emphasizing temporal change. The prominence of \emph{Continuity} is particularly notable, as it reflects the tendency of news discourse to frame events as part of ongoing trajectories, whether to legitimize claims via persistence, normalize repetition, or express dissatisfaction with an unchanging situation.

\paragraph{Co-occurrence of Frames}
Most temporally framed sentences contain a single frame (1,803 sentences), with progressively fewer sentences exhibiting multiple frames (468 with two frames, 86 with three, 7 with four, and 1 with five). To understand these cases, we examine which temporal frames appear together (see Appendix~\ref{sec:frame_cooccurrence}). \emph{Temporal Anchoring} co-occurs with \emph{Continuity}, \emph{Primacy}, and \emph{Temporal Contrast}. This is natural, as anchoring situates claims in time by reinforcing persistence, setting precedence via an anchor point, or highlighting change. \emph{Skeptical} framing shows moderate overlap with \emph{Urgency}, often in contexts where caution and doubt are raised about the future.

\paragraph{Corpus Splits}
To control for split-specific bias, we construct the train, development, and test splits such that each split contains a comparable proportion of documents with and without temporal framing \emph{within each language}. While the overall framing prevalence differs across languages, the splits preserve each language's internal framing distribution across documents. The dataset follows an approximate 70/13/17 train-development-test distribution. See Appendix~\ref{sec:corpus_splits} for a detailed breakdown across splits.

\paragraph{Lexical Analysis}
To characterize how temporal frames are realized at the surface level, we conduct a lexical association analysis comparing sentences annotated with each temporal frame against the rest of the corpus using smoothed log-odds ratios \citep{monroe2008fightin}. Overall, most frames exhibit intuitive lexical cues, while others rely less on stable surface forms. In particular, \emph{Temporal Anchoring} and \emph{Nostalgia} show weaker and more heterogeneous lexical signals resulting from their dependence on event-specific references and evaluative language rather than recurring temporal expressions. A detailed lexical analysis, including per-frame log-odds profiles and diagnostics, is provided in Appendix~\ref{sec:lexical_analysis}.

\begin{table*}[!ht]
\centering
\small
\setlength{\tabcolsep}{4pt}
\renewcommand{\arraystretch}{1.15}

\begin{tabular}{llccc|cccc}
\toprule
\textbf{Method} & \textbf{Model} & \multicolumn{3}{c}{\textbf{Binary Detection}} & \multicolumn{4}{c}{\textbf{Multilabel Classification}} \\
\cmidrule(lr){3-5} \cmidrule(lr){6-9}
 &  & P & R & F1 & P & R & Micro-F1 & Macro-F1 \\
\midrule
 & Random Baseline & 0.11 & 0.88 & 0.20 & 0.02 & 0.49 & 0.04 & 0.03 \\

\midrule
\multirow{6}{*}{Zero-Shot} & Llama-3.1-8B-Instruct & 0.14 & 0.94 & 0.24 & 0.06 & 0.34 & 0.10 & 0.11 \\
 & Llama-3.3-70B-Instruct & 0.28 & 0.78 & 0.41 & 0.16 & 0.43 & 0.23 & 0.21 \\ 
 & Qwen3-8B & 0.21 & 0.75 & 0.33 & 0.08 & 0.29 & 0.13 & 0.14 \\
 & Qwen3-32B & 0.36 & 0.57 & 0.44 & 0.20 & 0.31 & 0.25 & 0.22 \\
 & Qwen3-235B-A22B & 0.31 & 0.70 & 0.44 & 0.17 & 0.38 & 0.24 & 0.23 \\
 & GPT-5.2 & 0.34 & 0.67 & \textbf{0.45} & 0.24 & 0.46 & \textbf{0.31} & 0.29 \\

\midrule
\multirow{3}{*}{Supervised} & XLM-R & 0.38 & 0.75 & 0.51 & 0.28 & 0.57 & 0.37 & 0.33 \\
 & Llama-3.1-8B-Instruct & 0.67 & 0.45 & 0.54 & 0.48 & 0.37 & 0.42 & 0.35 \\
 & Qwen3-8B & 0.60 & 0.54 & \textbf{0.57} & 0.48 & 0.40 & \textbf{0.44} & 0.35 \\
\bottomrule
\end{tabular}
\caption{\textbf{Overall results on the test set.} Zero-shot models show high recall but low precision, while supervised models achieve more balanced performance with strong gains in detection and multilabel classification. The random baseline samples labels from the annotation label distributions in the combine training and development sets.}
\label{tab:overall_combined_table_star}
\end{table*}

\section{Experiments} 
\label{sec:experiments}


\subsection{Experimental Setup}

We evaluate two strategies for sentence-level temporal frame detection: (1) zero-shot classification and (2) supervised fine-tuning on our annotated corpus. The former tests out-of-the-box recognition without task-specific training, while the latter assesses learning under supervision, using a consistent input representation. This setup allows us to contrast implicit knowledge encoded in pretrained models with task-specific adaptation under controlled conditions.

We adopt a sentence-only input for all experiments, motivated by context ablation results in Appendix~\ref{appx:llm:context_ablation}. We also evaluate two extended variants: (i) neighboring context (preceding and following sentences) and (ii) full document context. Neither improves performance over sentence-only input and incurs higher token costs. Additional context makes models more conservative, often consistently leading to missed temporal frames in the target sentence during evaluation.

\paragraph{Zero-Shot Classification.}
We use both a proprietary LLM and open-weight language models of various scales for zero-shot inference. Specifically, we evaluate OpenAI’s GPT-5.2 alongside models from the LLaMA3 and Qwen3 architecture families. For the open-weight models, we consider sizes ranging from 8 to 235 billion parameters to examine the effect of scale on performance. We prompt all zero-shot models using a consistent instruction set (see Appendix~\ref{sec:appendix_zeroshot}), with additional inference and output-schema details provided in Appendix~\ref{appx:sec:llm_inference_details}.

\paragraph{Supervised Fine-Tuning.} 
For supervised fine-tuning, we consider both an efficient encoder-based model and selected language models from our zero-shot experiments. We adopt the pre-trained encoder model XLM-RoBERTa-base (XLM-R) \cite{conneau-etal-2020-unsupervised}, which has 270 million parameters, as an efficient baseline and train it as a multi-label sentence classification model. Given the extreme class imbalance in the data, where most sentences contain no temporal frame and only a small fraction exhibit multiple frames, we optimize a weighted binary cross-entropy loss that down-weights negative instances and apply label-aware batching to reduce batch-level sparsity and stabilize training (see Appendix~\ref{sec:training_details} for details). Beyond the encoder-based baseline, we fine-tune two representative language models from our zero-shot selection, namely Qwen3-8B and LLaMA3.1-8B. Additional details of the fine-tuning setup for these models are provided in Appendix~\ref{appx:sec:llm_finetuning_details}.






\begin{figure*}[t]
    \centering
    \includegraphics[width=0.7\textwidth]{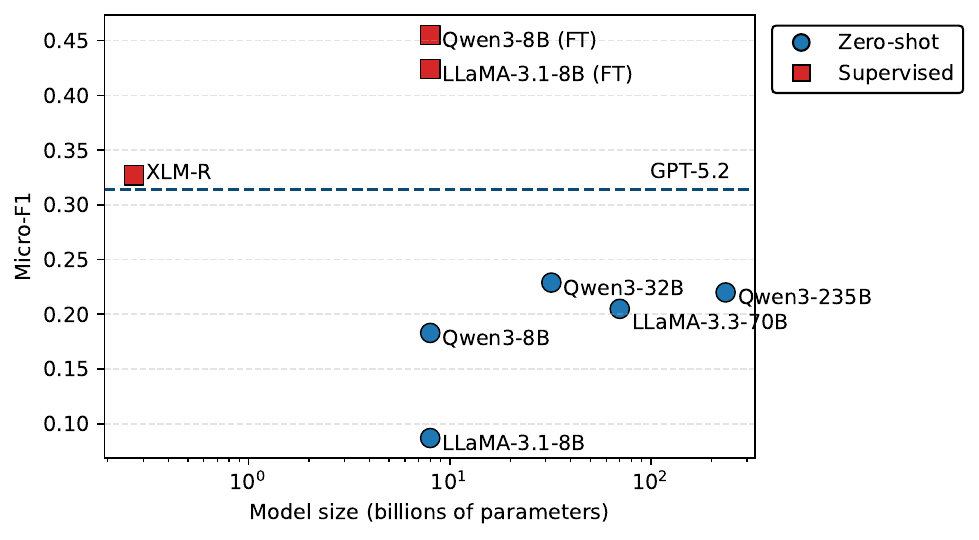}
    \caption{\textbf{Micro-F1 vs.\ model size on the test set.} Supervised fine-tuning (red squares) substantially outperforms zero-shot inference (blue circles) across all model scales. XLM-RoBERTa achieves competitive performance with only 0.27B parameters.
    }
    \label{fig:performance_vs_size}
\end{figure*}

\subsection{Results}
\label{sec:results}

Table~\ref{tab:overall_combined_table_star} summarizes results on the test set, which contains 3,517 sentences. We evaluate zero-shot configurations over ten runs and fine-tuned models over a single run, reporting mean scores across all metrics. We report both binary detection and multilabel classification performance. Figure~\ref{fig:performance_vs_size} visualizes Micro-F1 as a function of model size.

Among \emph{zero-shot approaches}, GPT-5.2 achieves the strongest performance, with a binary detection F1 of 0.45 and a multilabel Micro-F1 of 0.31. Within the Qwen family, binary F1 increases from 0.33 for Qwen3-8B to 0.44 for Qwen3-235B. 


A similar trend is observed for LLaMA models, where the 70B variant (0.41) substantially outperforms the 8B model (0.24), highlighting the effect of increased model capacity and representational power. Across zero-shot models, recall is generally high while precision remains low, indicating a tendency toward over-prediction and reduced reliability in practical settings and downstream applications. This suggests that, without task-specific supervision, LLMs struggle to reliably detect temporally framed sentences.


\emph{Supervision} yields substantial improvements over zero-shot baselines. XLM-R achieves a binary F1 of 0.51 and a Micro-F1 of 0.37 with only 270M parameters, which demonstrates that encoder-based models can effectively learn temporal framing patterns under direct supervision. For LLMs in the zero-shot setting, scale alone produces only modest gains, with improvements tapering as model size increases, suggesting diminishing returns from additional parameters and limited benefits from scaling alone (Figure~\ref{fig:performance_vs_size}). In contrast, fine-tuning leads to large performance gains: LLaMA-3.1-8B-Instruct improves by 125\% relative to its zero-shot counterpart, while fine-tuned Qwen3-8B achieves \emph{the best overall performance} with a binary F1 of 0.57 and a Micro-F1 of 0.44, corresponding to a 27\% relative improvement over the strongest zero-shot result. The fine-tuned LLMs also exhibit distinct precision--recall profiles: Qwen3-8B remains relatively balanced, whereas LLaMA-3.1-8B-Instruct is more conservative, achieving higher precision at the cost of lower recall across different evaluation scenarios and dataset conditions.

\subsection{Discussion}
\label{sec:experiments_commentary}

\paragraph{Learnability and scale limitations.} Our results (Figure~\ref{fig:performance_vs_size}) indicate that sentence-level temporal framing is learnable, but not reliably addressed by model scale alone in the zero-shot setting. While larger models yield incremental gains, performance saturates quickly, suggesting that temporal framing relies on subtle rhetorical cues and relational contrasts that benefit more from task-specific supervision than from additional parameters. This indicates that prediction errors are driven by label ambiguity, subtle rhetorical cues, and sparsity rather than raw model capacity, highlighting the importance of high-quality annotations and task-specific training signals. Supervised training consistently improves both binary detection and frame identification, supporting the view that temporal framing patterns are systematic but not easily recoverable through general-purpose prompting alone.



\paragraph{Frame-level behavior.} At a finer granularity, supplementary analysis in Appendix~\ref{appx:sec:llm_experiment_details} provides per-language and per-frame breakdowns that further contextualize these findings and reveal cross-lingual variation. Across zero-shot models, frames such as \emph{Continuity}, \emph{Temporal Contrast}, and \emph{Urgency} are more consistently detected, likely because they are more prevalent and associated with recurring lexical or structural cues (see Appendix~\ref{sec:lexical_analysis}). Supervised fine-tuning reduces this gap across most frames, improving robustness and consistency, though rare frames, particularly \emph{Nostalgia}, remain affected by data sparsity and limited training instances.

\paragraph{Cross-lingual and sparsity analysis.} Across languages, performance differs: per-language analysis shows lower performance on German data relative to English, which is expected given that the German subset contains substantially fewer framed sentences and fewer positive instances per frame, as discussed in Section~\ref{sec:statistics}. This sparsity disproportionately affects recall for low-frequency frames and contributes to greater variance across models, rather than indicating a failure of cross-lingual generalization. Future work that explicitly controls for sparsity and corpus composition, for example through targeted augmentation of rare frames, would help disentangle language from data distribution effects.

\paragraph{Model behavior and trade-offs.} Beyond these data effects, encoder-based and instruction-tuned models exhibit complementary strengths. XLM-R shows that a standard multilingual encoder, when trained with imbalance-aware objectives, can capture much of the underlying signal, achieving strong performance despite its relatively small size and surpassing all zero-shot baselines. In contrast, fine-tuned 8B LLMs deliver the strongest end-to-end performance under the same sentence-only input format. At the same time, they exhibit distinct precision--recall profiles: LLaMA-3.1-8B-Instruct is more conservative, favoring higher precision, whereas Qwen3-8B maintains a more balanced recall profile. These differences make the models better suited to different downstream settings, depending on whether minimizing false positives or maximizing coverage is the primary objective.

\section{Conclusion and Future Work}
\label{sec:conclusion}


We introduced \emph{temporal framing} as a distinct and computationally tractable dimension of persuasive language. We formalized temporal framing as the rhetorical use of time-related expressions to structure meaning rather than to report chronology, proposed a taxonomy of eight temporal frames grounded in social science theory, and realized it through a multilingual, sentence-level annotated news corpus in English and German. Our corpus analysis showed that temporal framing is frequent, heterogeneous, and often multi-faceted, with frames such as \emph{Continuity}, \emph{Temporal Anchoring}, and \emph{Temporal Contrast} playing a central role in news discourse and shaping how narratives evolve across different topics and contexts over time and across diverse media outlets.


Our experiments show that temporal framing is \emph{learnable at the sentence level}, but not reliably captured by model scale alone in zero-shot settings. Larger models yield only modest gains, with performance quickly saturating without supervision. In contrast, supervised fine-tuning substantially improves both detection and frame assignment, even for compact models, indicating that temporal framing patterns are systematic but require task-specific learning signals. Fine-tuned models also exhibit distinct precision--recall profiles: Qwen3-8B is more balanced, while LLaMA-3.1-8B-Instruct is more conservative, achieving higher precision at the cost of lower recall.

These findings showed that temporal framing introduced a level of abstraction beyond conventional temporal analysis in NLP. Unlike tasks focused on temporal expression extraction or event ordering, temporal framing required interpreting how time is used to guide interpretation and influence perception. This helped explain why zero-shot models struggled despite strong general language understanding, as the task involved recognizing rhetorical intent rather than surface patterns alone. At the same time, consistent gains from supervision indicated that these patterns are learnable when appropriate signals are provided.

This work provided a valuable resource and empirical foundation for studying the rhetorical functions of time in news discourse. While our work focused on sentence-level modeling, the corpus was designed to support broader analyses at multiple levels of discourse, including document-level analysis. Future work can build on this corpus to examine temporal framing across additional languages and domains, to model frame interactions beyond the sentence level, and to investigate how temporal framing shapes narrative structure at scale.

Finally, our results highlighted that temporal framing occupies a distinct space between surface-level temporal expression and higher-level narrative structure. While models can learn to detect framing at the sentence level, many cases require interpreting implicit rhetorical cues that extend beyond explicit temporal markers. Thus, future work should explore richer representations of discourse context, including cross-sentence dependencies and narrative progression, to better capture how temporal framing unfolds across longer text spans. Such extensions would enable a more complete account of how time is used not only to describe events, but to shape interpretation and persuasion.

\section*{Limitations}


\paragraph{Corpus}
The corpus comprises 458 English and German news articles from a fixed set of outlets. It is enriched for opinionated content to increase temporal framing coverage and is not representative of overall news production. The corpus is unbalanced across outlets and framing prevalence, with variation across languages. As it is limited to English and German news, findings may not generalize to other languages or media contexts.

\paragraph{Annotation Scope}
Annotations are performed at the sentence level rather than using span-level annotation, which may miss cases where temporal framing is realized within partial clauses or spans that do not align with sentence boundaries. In addition, annotations focus on authorial voice and exclude quoted and indirect speech, limiting coverage of framing conveyed through attribution.

\section*{Ethical Considerations}


\paragraph{Biases}
The corpus draws on a fixed set of news outlets and reflects their inherent biases. Although it includes diverse editorial orientations, it is not balanced across viewpoints or ideologies. Annotators were instructed to identify temporal framing independently of personal views or article stance.


\paragraph{Intended Use and Misuse Potential}
The corpus supports research on temporal framing and rhetorical analysis in news, aiding transparency and critical understanding of persuasive language. However, as with other framing resources, annotations could be misused to optimize persuasive or manipulative communication. The corpus is released for analytical purposes with awareness of this risk.


\paragraph{Annotation Practice and Fairness.}
Annotations were produced by trained annotators affiliated with the authors’ institutions, following detailed guidelines with periodic calibration. Annotators were compensated according to institutional standards, and no crowd-sourcing was used. The task does not assess factual accuracy, truthfulness, or political intent, and labels should not be interpreted as indicators of misinformation or bias.

\paragraph{Environmental Considerations}
This work makes use of large language models for inference. As with other LLM-based methods, this entails computational cost and contributes to energy consumption and associated carbon emissions.

\section*{Acknowledgments}

This work was partially supported by the German Federal Ministry for Research, Technology and Space (BMFTR) under projects ``VeraXtract'' (16IS24066) and ``news-polygraph'' (03RU2U151C).

\bibliography{consolidated,acl_anth}

\appendix
\clearpage
\section{Annotation Guidelines}
\label{sec:appendix}

The annotation for this task is done as follows:

\begin{enumerate}
    \item You should carefully read Section~\ref{sec:temporal_framing} before beginning annotation, in order to understand the definition of temporal framing, the taxonomy of frames, and key distinctions such as factual vs. opinionated use and general vs. temporal senses.

    \item You are provided with a number of news articles and are expected to identify sentences that contain temporal framing according to the taxonomy introduced in Figure~\ref{tab:temporal-framing-definitions}. 
    
    \item The taxonomy is provided in two forms:
    \begin{itemize}
        \item \textbf{Brief overview} (Figure~\ref{tab:temporal-framing-definitions}): concise definitions and examples for quick reference.
        \item \textbf{Detailed taxonomy description} (Section~\ref{sec:detailed_taxonomy}): extended definitions, rhetorical functions, positive as well as negative examples, and explanations for each example. The detailed description should be consulted for challenging and ambiguous cases.
    \end{itemize}

    \item You should \textbf{only mark positive instances of framing}. Negative examples in the taxonomy are included to illustrate potential pitfalls and help avoid incorrect annotations.

    \item A sentence should be annotated as a frame only if the \textbf{temporal element is central to its persuasive force}. If the temporal element is merely descriptive (factual reporting) or absent (non-temporal lookalike), then no frame should be assigned.

    \item You should rely only on the information in the text itself, not on external world knowledge or assumptions.

    \item When in doubt, apply the following heuristic: \emph{if removing the temporal expression leaves the persuasive force of the sentence unchanged, then it is not an instance of temporal framing.}
    For example, consider: ``The plan was announced last week and it is deeply flawed.''  
    \begin{itemize}
        \item With the temporal expression: the persuasive force comes from the claim that the plan is \emph{deeply flawed}.  
        \item Without the temporal expression (``The plan is deeply flawed''): the persuasive force remains identical.  
    \end{itemize}
    Thus, the temporal cue ``last week'' is incidental, and this sentence should not be considered for any temporal framing.
    
    \item Pay attention to common temporal cues, including but not limited to: 
    \textit{first, latest, now, then, still, again, since, until, before, after, within, deadline, turning point, every, seasonal, for years/decades/centuries, post\textendash9/11, pre-war, next year, last month.} Such cues may signal the existence of a temporal frame.

    \item Some sentences may include more than one temporal frame simultaneously (e.g., \emph{recency + urgency}). In such cases, you should assign all frames that apply.

    \item Context matters. A sentence that appears factual in isolation may still count as framing if the surrounding context utilizes its temporal component as a persuasive element.

    \item \textbf{Quoted or Indirect Speech Exclusion:}
Do not annotate any sentence  that is wholly within quotation marks or represents indirect or reported speech (e.g., ``The minister said the crisis will worsen next year''). 
Temporal framing analysis applies only to the narrator's or authorial voice, not to speech attributed to others. 

\textbf{Example:}
\begin{quote}
``Back then, I simply was not as educated as to the implications and consequences of my actions as I have become since,'' Knight said.
\end{quote}
\textit{Annotation: None: indirect or quoted speech is excluded from analysis.}
\newline
\textit{Note:} Had this not been a quotation, the contrast between ``back then'' and ``since'' would qualify as temporal contrast, as it juxtaposes a past state of ignorance with a present state of awareness.

\item \textbf{Cross-Sentence Temporal Framing}  
Temporal framing may rely on information introduced in earlier sentences.  
In such cases, \textbf{annotate only the sentence where the temporal framing is made explicit}, even if earlier sentences provide necessary context.

\textbf{Example:}
\begin{quote}
$S_{1}$: For decades, the agency was trapped in slow, paper-based workflows. \\
$S_{2}$: Today, it operates digitally, transforming weeks of delays into same-day approvals.
\end{quote}
\textit{Annotation:} $S_{2}$: \textbf{Temporal Contrast} (past vs.\ present).



\end{enumerate}

\paragraph{Recommended workflow.}
\begin{enumerate}
    \item Read the sentence carefully, keeping the surrounding context in mind, and note any temporal cues, both explicit (e.g., ``today,'' ``last year,'' ``within weeks'') and implicit (e.g., historical landmarks such as ``the Cold War era,'' ``Brooklyn's protests''). 

    \item \textbf{First check: Quoted/indirect speech.}
    If the sentence is wholly or primarily direct or indirect speech, do not annotate and skip. 
    Consider for annotation only if the author's/narrator's sentence independently carries the temporal framing.

    \item \textbf{Second check: Fact vs.\ rhetorical.}  
    Decide whether the temporal element is merely \emph{descriptive} (factual reporting) or functions as \emph{rhetorical framing} (adds persuasive force).  
    If the sentence appears factual in isolation but surrounding context makes the temporal element rhetorically persuasive, then treat it as temporal framing. 
    
    \item \textbf{Third check: General vs.\ temporal sense.}  
    If the language only resembles framing in a general sense (e.g., contrast) but does not involve time as a framing resource, treat it as a non-temporal lookalike and do not annotate.

    \item \textbf{Fourth check: Cross-sentence (anchor-only).}
    If the sentence contains an incomplete temporal cue (\emph{now, still, no longer, has since, used to}) or implies a then-now contrast, scan the context for potential anchors. 
    Annotate \emph{only} the sentence where the persuasive temporal force culminates as the \emph{anchor}. 
    If the force culminates in a neighboring sentence, skip the current one and annotate that neighbor as the anchor.
    
    \item If the instance passes all checks, select the most appropriate frame(s) from the taxonomy.  
    
    \item In cases where multiple frames apply simultaneously, assign all that are relevant.  



    \item Use the compact taxonomy table for quick reference; consult the detailed taxonomy descriptions if uncertain. 

    \item \textbf{Challenging examples.}  
    If an example remains ambiguous even after consulting the taxonomy, you should note it down and raise it to the language coordinator for resolution.

\end{enumerate}

\section{Detailed Taxonomy Description}
\label{sec:detailed_taxonomy}
In what follows, we present each temporal frame with a more detailed description, including its \emph{persuasive function}, and examples that separate persuasive uses from mere factual reports and from non-temporal framing.

\noindent\textit{Note: Examples illustrate how statements may function as temporal framing devices. In some cases, the same wording could be read as neutral description when taken in isolation, but operate as framing when used rhetorically in context. The labels here indicate the intended illustrative function rather than an absolute classification.}

\newcommand{\frameentry}[4]{
\subsection{#1}
\textbf{Definition:} \textit{#2}

\vspace{6pt}

\noindent\textbf{Rhetorical function:} \textit{#3}

\vspace{6pt}


#4
}

\frameentry{Primacy}
{Primacy framing assigns significance to temporal precedence, treating leading positions as inherently meaningful. \\The precedence may be absolute in time (e.g., ``the earliest case ever'' or ``first since the beginning of the record'') or relative to a bounded interval (e.g., ``the earliest this year,'' ``first in this decade,'' or ``first under this administration'').}
{It converts the mere chronology of being first in time into persuasive force, establishing authority, legitimacy, or caution.}
{
\begin{itemize}

\item[\posicon] 
\textbf{``As the first city to adopt this policy, we set the standard for the region.''}  
\\[2pt]
\textit{Being first is explicitly linked to leadership and legitimacy, making temporal precedence the persuasive force.}



\vspace{6pt}

\item[\posicon] 
\textbf{``The collapse of the bank unsettled confidence in the markets to unprecedented levels.''}  
\\[2pt]
\textit{Low market confidence is framed as without precedent, amplifying the perceived danger of the bank’s collapse.}

\vspace{6pt}

\item[\negicon] 
\textbf{``The vaccine was approved first.''}  
\\[2pt]
\textit{Taken in isolation, this is a chronological report; it does not claim that being first entails special significance. However, if being first plays a persuasive role when the surrounding context is taken into account, then the same sentence could function as primacy framing.}

\vspace{6pt}

\item[\negicon] 
\textbf{``The best-performing vaccine won praise.''}  
\\[2pt]
\textit{The praise is grounded in quality (performance), not temporal primacy; it resembles primacy in a general sense but lacks the temporal element.}

\end{itemize}
}

\frameentry{Recency}
{Recency framing attributes significance to temporal proximity, treating the latest events as inherently meaningful. \\The proximity may be absolute in time (e.g., ``the most recent ever,'' ``latest on record''), relative to a bounded interval (e.g., ``latest this year,'' ``most recent this quarter,'' ``latest under this administration''), or anchored to a salient event (e.g., ``just hours after the event'' ``a few days after the ceasefire'') where closeness to that event is presented as meaningful.}
{It seizes attention through temporal proximity, subtly displaces older evidence without needing to disprove it, or pressures audiences to accept arguments as timely and relevant.}
{

\begin{itemize}
    \item[\posicon]  \textbf{``Today’s figures matter most for judging performance.''}   \\[2pt]  
    \quad \textit{Freshness is explicitly framed as the reason for relevance.}
    
    \item[\posicon]  \textbf{``Just released footage shows what really happened.''}   \\[2pt]  
    \quad \textit{The recency of the footage is used as proof of accuracy and priority.}
    
    \item[\negicon]  \textbf{``A recent poll shows that 60\% of voters support the policy.''}   \\[2pt]  
\quad \textit{Although temporally recent, this is straightforward factual reporting rather than a persuasive appeal to recency.}

    \item[\negicon]  \textbf{``In a recent speech, a respected senator criticized the plan and rightfully so.''}   \\[2pt]  
\quad \textit{Although it mentions recency, the persuasive force comes from the senator’s authority and opinion rather than the timing. The temporal marker is incidental, not persuasive.}

\end{itemize}
}

\frameentry{Urgency}
{Urgency framing creates a sense of immediacy or imminence by emphasizes limited time, decisive moments, ticking clocks, last chances, ultimatums, or imminent threats.}
{It depicts situations as temporally constrained or approaching a critical moment in order to heighten perceived stakes and pressure, or call for immediate action.}
{
\begin{itemize}
    \item[\posicon]  \textbf{``Congress must act within 72 hours to prevent a government shutdown..''}   \\[2pt]  
    \quad \textit{A specific deadline creates immediacy and pressure for political decision-making.}
    
    \item[\posicon]  \textbf{``This election is the defining moment for our nation’s future..''}   \\[2pt]  
    \quad \textit{Frames the present moment as uniquely decisive and urgent.}
    
    \item[\negicon]  \textbf{``Most voters say healthcare is an important issue..''}   \\[2pt]  
    \quad \textit{Conveys significance, but without compressed time or urgency.}
    
    \item[\negicon]  \textbf{``Only a limited number of senators support the bill..''}   \\[2pt]  
    \quad \textit{Indicates scarcity of support, but not temporal scarcity; no deadline or ticking clock.}
\end{itemize}
}

\frameentry{Temporal Anchoring}
{Temporal anchoring frames the discussion through the lens of past events that serve as interpretive anchors.}
{{It activates shared memory and emotional resonance, using familiar reference points to legitimize current perspectives or priorities, or to bias interpretation through invocation of certain events in the past.}
}

{
\begin{itemize}
    \item[\posicon]  \textbf{``We live in a post-9/11 world where national security must come first, and the congress must take immediate action.''}   \\[2pt]  
    \quad \textit{The landmark date frames a lasting shift in values and policy priorities.}
    
    \item[\posicon]  \textbf{``This legislation represents the boldest reform since the New Deal.''}   \\[2pt]  
    \quad \textit{Frames present significance by situating it relative to a landmark era.}
    
    \item[\negicon]  \textbf{``Security policies were tightened after 9/11.''}   \\[2pt]  
    \quad \textit{Merely states historical sequence.}
    
    \item[\negicon]  \textbf{``We live in a world where national security must come first, and the congress must take immediate action.''}   \\[2pt]  
    \quad \textit{It is the same example from earlier, but the omission of the temporal reference to 9/11. So while it is persuasive, it is not anchored in time.}
\end{itemize}
}

\frameentry{Nostalgia}
{Nostalgia framing invokes a cherished past as an ideal or standard for the present or future. }
{It activates a sense of identity and belonging, calls for restoration and frames it as progress, and softens resistance to a familiar past.}
{
\begin{itemize}
\item[\posicon]  \textbf{``We must return to the prosperity of the postwar years.''}   \\[2pt]  
\quad \textit{Frames a remembered era of economic success as the standard for present policy.}
    
    \item[\posicon]  \textbf{``We need to revive the spirit of bipartisanship that once defined Congress.''}   \\[2pt]  
    \quad \textit{Invokes a remembered political culture as guidance for present reform.}
    
    \item[\negicon]  \textbf{``The city is restoring historical facades downtown.''}   \\[2pt]  
    \quad \textit{Reports an action but does not frame the past as a persuasive political ideal.}
    
    \item[\negicon]  \textbf{``Many voters say they value tradition.''}   \\[2pt]  
    \quad \textit{Expresses a general liking for continuity, but without invoking a shared past as a model for today.}
\end{itemize}
}

\frameentry{Temporal Contrast}
{Temporal contrast framing emphasizes change by juxtaposing different time periods such as ``then'' versus ``now.''}
{It dramatizes contrast to mark decline, progress, or reversal, thereby justifying action or interpretation.}

\begin{itemize}
    \item[\posicon]  \textbf{``Once a neglected district, now a thriving hub.''}   \\[2pt]  
    \quad \textit{Direct before--after comparison makes temporal change salient.}
    
    \item[\posicon]  \textbf{``Twenty years ago tuition was manageable, today it traps students in debt.''}   \\[2pt]  
    \quad \textit{Uses then vs.\ now contrast to highlight deterioration.}
    
    \item[\negicon]  \textbf{``Contrasting 2005 to 2010, the downtown population increased by 20\%.''}   \\[2pt]  
    \quad \textit{Without full context, this sentence is a factual report of a statistical change without rhetorical contrast.}
    
\item[\negicon]  \textbf{``Urban elites and rural voters have conflicting values, and policymakers must bridge this divide.''}   \\[2pt]  
    \quad \textit{Makes a rhetorical appeal based on geographical contrast, not a temporal one.}
\end{itemize}

\frameentry{Continuity}
{Continuity framing expresses persistence across time, encompassing stability, stagnation, momentum, or recurrence.}
{It legitimizes through longevity, reassures through endurance, or normalizes repetition by presenting ongoing trajectories or cycles as natural and inevitable. It can also convey discontent, dissatisfaction, or frustration with an unchanging situation.}

{
\begin{itemize}
    \item[\posicon]  \textbf{``For centuries this constitution has safeguarded our liberties..''}   \\[2pt]  
    \quad \textit{Longevity is invoked as a direct source of legitimacy, even though whether the constitution truly safeguarded liberties is a matter of interpretation and debate.}

    \item[\posicon]  \textbf{``Our neutrality has endured for generations, and it continues to serve us well..''}   \\[2pt]  
    \quad \textit{Persistence over time is framed as proof of value and stability.}

\item[\posicon] 
\textbf{``Every election season, the same promises return.''}  
\\[2pt]
\textit{Uses recurrence as both explanation and expression of discontent and critique.}

\vspace{6pt}

\item[\posicon] 
\textbf{``Markets crash and recover; this downturn will pass.''}  
\\[2pt]
\textit{Frames recurrence as reassurance.}

\vspace{6pt}

\item[\posicon] 
\textbf{``Our coalition has grown election after election, and nothing can stop this progress now.''}  
\\[2pt]
\textit{Past trajectory is invoked to suggest inevitability.}

\vspace{6pt}

\item[\posicon] 
\textbf{``This campaign gains strength with every new supporter who joins us.''}  
\\[2pt]
\textit{Accumulating participation is framed as persuasive momentum.}

\vspace{10pt}

    \item[\negicon]  \textbf{``The law remains on the books.''}   \\[2pt]  
    \quad \textit{Factually reports longevity, but without using it in a persuasive way.}
    
\item[\negicon]  \textbf{``The 1990 law is widely supported by voters.''}   \\[2pt]  
\quad \textit{Legitimacy is attributed to present popularity, not to persistence or longevity over time.}

\item[\negicon] 
\textbf{``As every winter, flu cases surge again.''}  
\\[2pt]
\textit{Observes a seasonal pattern but does not employ it rhetorically.}

\vspace{6pt}

\item[\negicon] 
\textbf{``The flu is a serious threat to public health and cannot be ignored.''}  
\\[2pt]
\textit{Makes an evaluative claim without invoking recurrence.}

\vspace{6pt}

\item[\negicon] 
\textbf{``The rally had great energy and everyone was fired up.''}  
\\[2pt]
\textit{Shows force and enthusiasm in the moment,
but not a trajectory over time.}

\vspace{6pt}

\item[\negicon] 
\textbf{``Voter turnout has increased in each of the last three elections.''}  
\\[2pt]
\textit{Without the full context, this sentence merely reports momentum, but without any rhetorical appeals.}

\end{itemize}
}

\frameentry{Skeptical}
{Skeptical framing casts doubt about the future  by emphasizing uncertainty, risk, pessimism, potential failure, or negative outcomes}
{It highlights vulnerability or doubt to provoke caution, hesitation, or reassessment of confidence in projected outcomes.}

{
\begin{itemize}
    \item[\posicon]  \textbf{``The plan may collapse under financial pressure..''}   \\[2pt]  
    \quad \textit{Draws attention to risk in order to slow or block action.}
    
    \item[\posicon]  \textbf{``The merger could backfire on consumers..''}   \\[2pt]  
    \quad \textit{Frames a potential outcome as harmful and uncertain.}
    
    \item[\negicon]  \textbf{``The plan costs 2 billion dollars..''}   \\[2pt]  
    \quad \textit{A financial fact, not a speculative projection.}
    
    \item[\negicon]  \textbf{``As history shows, a ceasefire will eventually be signed..''}   \\[2pt]  
    \quad \textit{Treats the outcome as inevitable certainty rather than a contingent risk.}
\end{itemize}
}

\section{Additional Corpus Analyses}
\label{sec:appendix_corpus_analysis}

\subsection{Unsupervised Document Tagging Reliability}
\label{sec:llm_tagging_reliability}

As described in Section~\ref{sec:article_selection}, we used LLM-assisted document-level opinion tagging to guide sampling, with the goal of enriching the dataset with articles likely to contain persuasive temporal framing. Prior to sampling, only 22.1\% of English articles and 13.4\% of German articles were tagged as opinionated. 
This indicated that such content formed a minority of the raw corpus. To ensure sufficient coverage, we upsampled documents tagged as opinionated while still retaining a substantial subset of non-opinionated ones to account for potential noise in these LLM generated tags.

After sampling, non-opinionated articles constituted 28.6\% of the English subset and 25.5\% of the German subset, resulting in final opinion-tag rates of 71.4\% and 74.5\%, respectively.

We now evaluate the reliability of this weak document-level signal by comparing LLM opinion tags against sentence-level human temporal framing annotations.

Using the presence of at least one temporally framed sentence as ground truth, 84.7\% of documents tagged as opinionated by the LLM contained temporal framing, corresponding to a high true-positive rate for detecting framing-bearing articles. 

In contrast, 41.9\% of documents tagged as non-opinionated still exhibited temporal framing, indicating a substantial false-negative rate when opinionatedness is used as a proxy for framing presence. 

These results confirm that LLM-based opinion tagging is strongly correlated with temporal framing, but does not fully capture its distribution.

Beyond binary presence, documents tagged as opinionated also exhibited higher framing intensity. Across all documents, LLM-opinionated articles showed a higher mean ratio of framed sentences (0.150 vs.\ 0.067) and a higher mean number of temporal frame assignments per sentence (0.194 vs.\ 0.085) compared to LLM-non-opinionated articles. When restricting the comparison to documents containing at least one framed sentence, opinionated articles still displayed slightly higher framing density (mean framed-sentence ratio 0.177 vs.\ 0.161) and substantially more framed sentences per document on average (7.55 vs.\ 4.37). 

These findings corroborate the soundness of our sampling strategy: while LLM opinion tags were effective for enriching the dataset with framing-heavy documents, retaining a controlled proportion of LLM-negative articles was necessary to capture temporal framing in ostensibly factual reporting. LLM predictions were used exclusively for document-level sampling decisions; all temporal framing labels and intensity measures were derived solely from human annotations.

\subsection{Framewise Inter-Annotator Agreement at the Sentence-Level}
\label{sec:iaa_sent_level_framewise}

Table~\ref{tab:iaa_per_frame} reports sentence-level inter-annotator agreement scores computed separately for each temporal frame. Agreement is measured using Krippendorff's $\alpha$ (nominal), treating each frame as a binary label (present vs.\ absent). These framewise scores complement the aggregate agreement results reported earlier in Table~\ref{tab:iaa_summary} and illustrate variability in annotation difficulty across temporal frames.

\begin{table}[h]
\centering
\small
\setlength{\tabcolsep}{6pt}
\renewcommand{\arraystretch}{1.15}
\begin{tabular}{lcc}
\toprule
Temporal Frame & $\alpha_\text{English}$ & $\alpha_\text{German}$ \\
\midrule
Primacy            & 0.506 & 0.768 \\
Recency            & 0.469 & 0.302 \\
Urgency            & 0.442 & 0.593 \\
Temporal Anchoring & 0.590 & 0.434 \\
Nostalgia          & 0.416 & 0.666 \\
Temporal Contrast  & 0.538 & 0.640 \\
Continuity         & 0.552 & 0.630 \\
Skeptical          & 0.443 & 0.674 \\
\bottomrule
\end{tabular}
\caption{\textbf{Sentence-level per-frame inter-annotator agreement.} It is measured with Krippendorff's $\alpha$ (nominal), treating each temporal frame as a binary label (present vs.\ absent).}
\label{tab:iaa_per_frame}
\end{table}

\subsection{Inter-Annotator Agreement at the Document-Level}
\label{sec:iaa_doc_level}

To compute document-level labels, we take, for each annotator, the union of sentence-level frame assignments within an article. At the document level (Tables~\ref{tab:doc_iaa_summary}--\ref{tab:doc_iaa_per_frame}), agreement increases, which indicates that annotators agree more on overall framing than on sentence boundaries. 
Document-level detection achieves $\alpha=0.679$ (EN) and $\alpha=0.802$ (DE), while mean per-frame agreement increases to 0.579 (EN) and 0.692 (DE). Document-level set agreement remains high, with Set-F1$_{+}=0.761$ for English and 0.710 for German. Finally, document-level framing intensity in Table~\ref{tab:doc_intensity} shows rank agreement across annotators, with Spearman correlations of 0.575 and 0.541 (EN) and 0.839 and 0.805 (DE) for framed-sentence and frame-assignment counts, respectively.

\begin{table}[h]
\centering
\small
\setlength{\tabcolsep}{4pt}
\renewcommand{\arraystretch}{1.15}
\begin{tabular}{lrrrrrr}
\toprule
Lang. & $N_{\text{docs}}$ & $\alpha_{\text{det}}$ & $\overline{\alpha}_{\text{frame}}$ & min & max & Set-F1$_{+}$ \\
\midrule
English & 50 & 0.679 & 0.579 & 0.431 & 0.702 & 0.761 \\
German & 51 & 0.802 & 0.692 & 0.405 & 0.880 & 0.710 \\
\bottomrule
\end{tabular}
\caption{\textbf{Document-level inter-annotator agreement.} Document-level labels are obtained by taking, for each annotator, the union of sentence-level temporal frame assignments within an article. Metrics are defined analogously to the sentence-level setting.}
\label{tab:doc_iaa_summary}
\end{table}

\begin{table}[h]
\centering
\small
\setlength{\tabcolsep}{6pt}
\renewcommand{\arraystretch}{1.15}
\begin{tabular}{lcc}
\toprule
Temporal Frame & $\alpha_\text{English}$ &  $\alpha_\text{German}$  \\
\midrule
Primacy            & 0.620 & 0.880 \\
Recency            & 0.467 & 0.405 \\
Urgency            & 0.658 & 0.642 \\
Temporal Anchoring & 0.432 & 0.623 \\
Nostalgia          & 0.679 & 0.652 \\
Temporal Contrast  & 0.702 & 0.680 \\
Continuity         & 0.638 & 0.795 \\
Skeptical          & 0.431 & 0.863 \\
\bottomrule
\end{tabular}
\caption{\textbf{Document-level per-frame inter-annotator agreement.} It is measured with Krippendorff's $\alpha$ (nominal), using document-level frame presence derived from sentence-level annotations.}
\label{tab:doc_iaa_per_frame}
\end{table}

\noindent
\begin{minipage}{\linewidth}
\centering
\small
\setlength{\tabcolsep}{3pt}
\renewcommand{\arraystretch}{1.15}
\begin{tabular}{lrrrr}
\toprule
Lang. & $\rho_{\text{sent}}$ & MAE$_{\text{sent}}$ & $\rho_{\text{frame}}$ & MAE$_{\text{frame}}$ \\
\midrule
English & 0.575 & 5.67 & 0.541 & 7.93 \\
German & 0.839 & 2.43 & 0.805 & 3.78 \\
\bottomrule
\end{tabular}

\vspace{2pt}
\captionof{table}{\textbf{Document-level agreement on temporal framing intensity.} $\rho$ denotes Spearman rank correlation and MAE the mean absolute error. Subscripts indicate whether scores are computed over framed sentences or frame assignment counts.}
\label{tab:doc_intensity}
\end{minipage}

\subsection{Temporal Frame Co-occurrence}
\label{sec:frame_cooccurrence}
\begin{figure}[h]
    \centering
    \includegraphics[width=1\linewidth]{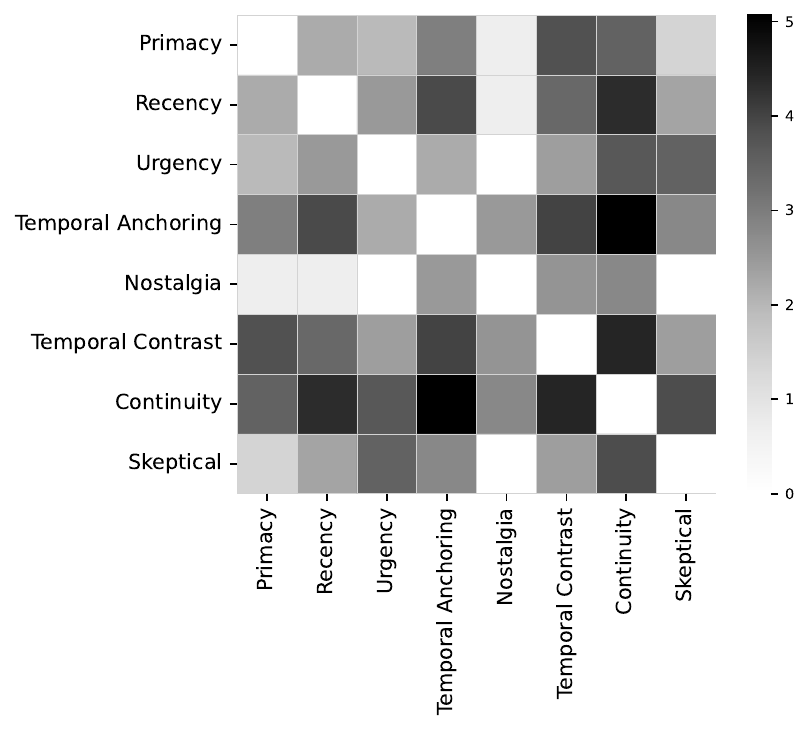}
    \caption{\textbf{Log-scaled heatmap of co-occurrence of temporal frames in the corpus.} Darker cells indicate more frequent co-occurrence patterns.}
    \label{fig:frames_cooccurence_heatmap}
\end{figure}

Figure~\ref{fig:frames_cooccurence_heatmap} shows the sentence-level co-occurrence patterns of temporal frames in the corpus. Values are log-scaled to highlight relative associations between frame pairs and illustrate the multi-label nature of temporal framing.

\subsection{Corpus Splits}
\label{sec:corpus_splits}
Table~\ref{tab:dataset_split} reports the distribution of documents across the train, development, and test splits for each language, including the number and proportion of documents containing at least one temporally framed sentence. The splits were constructed to preserve, within each language, the relative prevalence of temporally framed and non-framed documents, while maintaining an approximate 70/13/17 train-development-test ratio. This stratification reduces split-specific bias and ensures comparable framing distributions across splits. As a result, models are evaluated under consistent conditions that reflect the underlying corpus composition.

\begin{table}[h]
\centering
\small
\setlength{\tabcolsep}{4pt}
\renewcommand{\arraystretch}{1.15}
\resizebox{\columnwidth}{!}{%
\begin{tabular}{lrrr|r}
\toprule
 & Train & Dev & Test & All \\
\midrule
English & 138/166 \textcolor{black!60}{(83.1\%)} & 27/33 \textcolor{black!60}{(81.8\%)} & 33/39 \textcolor{black!60}{(84.6\%)} & 238 \\
German & 95/153 \textcolor{black!60}{(62.1\%)} & 18/30 \textcolor{black!60}{(60.0\%)} & 24/37 \textcolor{black!60}{(64.9\%)} & 220 \\ \hline
Total & 233/319 \textcolor{black!60}{(73.0\%)} & 45/63 \textcolor{black!60}{(71.4\%)} & 57/76 \textcolor{black!60}{(75.0\%)} & 458 \\
\bottomrule
\end{tabular}%
}
\caption{\textbf{Corpus distribution across train, dev, and test splits.}
Entries show \emph{framed/total} documents per split (percentage in parentheses); for example, \emph{138/166 (83.1\%)} indicates that 138 of 166 English training articles contain at least one temporal framing instance.}
\label{tab:dataset_split}
\end{table}

\subsection{Lexical Analysis}
\label{sec:lexical_analysis}

To better understand how temporal framing is realized at the surface level, we conduct a lexical analysis of sentences annotated with temporal frames. Rather than defining frames purely in lexical terms, this analysis aims to identify words that are disproportionately represented in sentences annotated with a given frame relative to the corpus. We quantify these associations using smoothed log-odds ratios \citep{monroe2008fightin}, which assign higher scores to words that occur more frequently in sentences expressing a given frame than expected based on their overall corpus frequency.
\begin{table}[h]
\centering
\small
\setlength{\tabcolsep}{3pt}
\renewcommand{\arraystretch}{1.15}
\begin{tabularx}{\columnwidth}{lX}
\toprule
Temporal Frame & Most Associated Terms \\
\midrule

Primacy &
\emph{record}, \emph{arrivals}, \emph{first}, \emph{ever}, \emph{democratic}, \emph{lot}, \emph{never}, \emph{history}, \emph{ufc}, \emph{high}, \emph{year}, \emph{since} \\
\midrule

Recency &
\emph{recently}, \emph{latest}, \emph{recent}, \emph{navy}, \emph{southern}, \emph{hours}, \emph{weeks}, \emph{week}, \emph{tariffs}, \emph{friday}, \emph{brought}, \emph{study} \\
\midrule

Urgency &
\emph{clinicians}, \emph{essential}, \emph{medical}, \emph{must}, \emph{soon}, \emph{threat}, \emph{rise}, \emph{citizens}, \emph{critical}, \emph{needs}, \emph{military}, \emph{today} \\
\midrule

Temporal Anchoring &
\emph{neocon}, \emph{battle}, \emph{copy}, \emph{century}, \emph{civil}, \emph{refused}, \emph{term}, \emph{harry}, \emph{pandemic}, \emph{rural}, \emph{reminder}, \emph{remained} \\
\midrule

Nostalgia &
\emph{you}, \emph{have} \\
\midrule

Temporal Contrast &
\emph{previous}, \emph{longer}, \emph{amendment}, \emph{ago}, \emph{plastic}, \emph{now}, \emph{once}, \emph{become}, \emph{today}, \emph{until}, \emph{research} \\
\midrule

Continuity &
\emph{steady}, \emph{increasingly}, \emph{repeatedly}, \emph{streak}, \emph{spirit}, \emph{methods}, \emph{scandal}, \emph{remained}, \emph{cryptocurrency}, \emph{continues}, \emph{remains}, \emph{volatile} \\
\midrule

Skeptical &
\emph{non-citizens}, \emph{spouses}, \emph{affect}, \emph{non-citizen}, \emph{qdot}, \emph{estate}, \emph{laws}, \emph{spouse}, \emph{senators}, \emph{planning}, \emph{broader}, \emph{tariffs} \\
\bottomrule
\end{tabularx}
\caption{\textbf{Top terms most strongly associated with each temporal frame, estimated using log-odds keyness.} Words are shown in descending order of association strength.}
\label{tab:lexical-logodds}
\end{table}

The resulting lexical profiles in Table~\ref{tab:lexical-logodds} summarize the most strongly associated words for each frame and reveal that most frames exhibit clear and intuitive surface realizations. For example, Recency is strongly associated with words such as \emph{recently}, \emph{today}, and \emph{latest}, while Urgency is characterized by markers of necessity and imminence such as \emph{must}, \emph{critical}, and \emph{soon}. Temporal Contrast similarly shows distinctive cues related to comparison across time, including \emph{previous}, \emph{once}, and \emph{now}.

However, Temporal Anchoring, Skeptical, and Nostalgia frames exhibit comparatively weaker lexical signals in the log-odds analysis. For Temporal Anchoring, this outcome is expected given its primary function of situating claims with respect to specific events, periods, or historical reference points. As a result, anchoring expressions naturally co-occur with event- or domain-specific vocabulary, rather than drawing from a small, recurring set of surface cues. This highlights the dependence of anchoring on contextual grounding rather than fixed lexical markers.

For Skeptical framing, we complement log-odds with a lexicon-based diagnostic targeting hedging and uncertainty terms. Such cues occur in 49.2\% of Skeptical-labeled sentences, compared to 10.8\% elsewhere in the corpus (a 4.56$\times$ enrichment). This signal is driven primarily by modal verbs expressing contingency (e.g., \emph{could}, \emph{would}, \emph{may}, \emph{might}) as well as explicit uncertainty terms (e.g., \emph{risk}, \emph{concern}, \emph{uncertainty}). This indicates that Skeptical framing is lexically realized through uncertainty and qualification rather than topical content.

Nostalgia framing in our corpus is primarily realized through evaluative and commemorative language rather than explicit temporal expressions. Nostalgia-labeled sentences frequently invoke idealized past states, historical figures, or perceived loss (e.g., references to legacy, vanished glory, or return), often without overt temporal adverbials such as \emph{then} or \emph{back}. Consequently, simple temporal cue-based analysis tends to under represent Nostalgia. Notably, the log-odds analysis for Nostalgia shows words such as \emph{you} and \emph{have}, which reflect the topical prevalence of commemorative and obituary-style writing within Nostalgia-labeled instances rather than nostalgia-specific lexical cues.

\section{Social Science Roots of the Temporal Framing Taxonomy}
\label{sec:social_science_roots}


In this section, we ground each temporal frame from our taxonomy to foundational theories from psychology, cognitive science, communication, sociology, and political science. These theoretical underpinnings shed light on why certain temporal frames resonate so strongly with audiences, based on how people process time-related information in memory, attention, and decision-making. We further illustrate how such concepts map onto the frame's persuasive function in discourse and communication across domains and contexts.

\subsection{Primacy -- First Impressions and Anchoring}

The \emph{Primacy} frame, which attributes special significance to being first in time, leverages one of the most established cognitive biases: the primacy effect. In psychology, the primacy effect refers to the tendency for information encountered earlier to be weighted more heavily than information encountered later~\cite{asch1946forming}. Early inputs receive greater attention, making them more likely to be encoded into long-term memory~\cite{serialpositioneffect1962,murdock1962serial}. Once an initial interpretive framework is established, later details are often assimilated into that framework rather than fundamentally revising it.

Classic work by \citet{asch1946forming} demonstrates that describing an individual with positive traits first leads to a more favorable overall impression than presenting the same traits in reverse order. In the news context, a Primacy frame can set the tone by emphasizing a first event, precedent, or milestone, thereby adjusting audience expectations for everything that follows. This connects closely to the anchoring heuristic in decision-making, where initial information serves as a mental benchmark that biases subsequent judgments~\cite{tversky1974judgment,prospect_theory,furnham2011literature}. By highlighting the first discovery, first warning, or unprecedented event, communicators tap into an audience's cognitive bias to favor the starting point in a sequence. Primacy framing draws on a robust cognitive tendency whereby what comes first enjoys a durable advantage in memory, interpretation, and influence.

\subsection{Recency -- Latest Events and the Availability Heuristic}

The \emph{Recency} frame emphasizes the most recent developments and is grounded in well-established findings from memory research~\cite{serialpositioneffect1962,murdock1962serial}. The recency effect describes the improved recall of information presented most recently, particularly when no substantial delay intervenes. Beyond recall, recent events exert disproportionate influence through the availability heuristic~\cite{TVERSKY1973207}, whereby people estimate importance, likelihood, or frequency based on how easily examples come to mind. As a result, recently reported events are often perceived as more representative of underlying trends than older evidence. 

Because recent or vivid incidents are more accessible in memory, they are often perceived as more important or representative than older information, which can distort risk perception. For example, a surge of recent crime reports may lead people to believe crime is increasing even when long-term trends show stability or decline. The Recency frame capitalizes on this tendency by foregrounding today's figures or latest developments as the most relevant indicators of reality. In continuous news cycles, the newest headline often displaces earlier context, reinforcing significance through recency. Overall, this frame reflects our tendency to treat recent events as especially informative.

\subsection{Urgency -- Imminence, Time Pressure, and Present Bias}

The \emph{Urgency} frame highlights limited time or imminent consequences and draws on both classical rhetoric and contemporary behavioral science. In ancient Greek rhetorical theory, the distinction between \emph{chronos} and \emph{kairos} provides a foundational account of how time functions persuasively \cite{smith1969time}. \emph{Chronos} refers to chronological, sequential time measured in objective units such as dates, durations, or intervals. By contrast, \emph{kairos} denotes the qualitatively ``right'' or opportune moment for action, which emphasizes timeliness, and consequence rather than metric placement on a timeline.

Kairotic appeals frame moments as fleeting, critical, or irrevocable, creating pressure that can be highly motivating. While references to chronos may appear on the surface, such as deadlines or countdowns, the persuasive force of urgency arises when these are rhetorically transformed into kairotic claims about actionability. The Urgency frame thus operates primarily through \emph{kairos} rather than \emph{chronos}, using time as a rhetorical device rather than a neutral dimension.

In psychology, this logic is echoed in scarcity principles~\cite{cialdini2001science}. When opportunities or resources are framed as time-limited, they trigger fear of missing out and prompt rapid decision-making. From a behavioral economics perspective, the effectiveness of Urgency is closely tied to present bias~\cite{ainslie1975specious,o2015present}, which describes the tendency to give more significance to immediate rewards or threats. People prefer smaller rewards now over larger rewards later, and they react more to dangers portrayed as imminent than to those framed as distant~\cite{trope2010construal}.

By compressing the temporal horizon, the Urgency frame counteracts temporal discounting, which makes future consequences feel concrete and immediate. By emphasizing that the next hours or days are critical focuses attention, it heightens the sense of alert, and encourages action. Whether framed through kairos or present bias, by making time scarce, one reliably intensifies engagement and motivates immediate response.

\subsection{Temporal Anchoring -- Contextualizing the Present through the Lens of the Past}

\emph{Temporal Anchoring} contextualizes present events by anchoring them to past moments. This strategy mirrors the anchoring effect identified by~\citet{tversky1974judgment}, whereby judgments are disproportionately influenced by an initial reference point. In the temporal domain, a past event could serves as a powerful anchor that shapes interpretation of current developments.

Once an anchor is established, subsequent information is evaluated relative to it. Referring to a contemporary issue as occurring in a ``post-9/11 world,'' for instance, activates a rich schema associated with that historical event, and it guides interpretation and emotional response. Schema theory~\cite{carbon2012bartlett} helps us understand that when one anchors to past events, such anchoring activate networks of associated beliefs that structure understanding of new information.

In political communication and sociology, historical analogies are widely used to legitimize or critique present actions~\cite{wodak2015politics}. By anchoring to significant moments in the past such as past crises, or previous conflicts, one transfers the schematic memory from that past to the present situation. Temporal anchors are often emotionally charged, drawing on wars, disasters, or triumphs to imbue current issues with significance. The persuasive power of this frame lies in our reliance on the past as a reference point for making sense of the present. By establishing history as an anchor, communicators guide audiences toward a preferred interpretation of ongoing events. The persuasive power of this frame lies in our reliance on the past as a reference point for making sense of the present. By establishing history as an anchor, communicators guide audiences toward a preferred interpretation of ongoing events. This makes temporal anchoring a powerful tool for shaping narrative coherence and meaning.

\subsection{Nostalgia -- Rose-Tinted Memory and Positive Past Identity}

The \emph{Nostalgia} frame invokes an idealized past and resonates strongly due to the reconstructive nature of human memory~\cite{Starobinski_1966,MITCHELL1997421}. Nostalgia is not simple recollection of past events but an emotion-driven reconstruction in which negative details are often downplayed and positive aspects are amplified. This reconstructive bias produces a rosy, idealized image of earlier times, which makes them appear more appealing than the present.

Social psychological research shows that nostalgia serves important identity functions. It induces a sense of self-continuity across time by linking present identity to valued aspects of the past~\cite{nostalgia2008}. By recalling where individuals or groups came from, nostalgia reinforces the idea of a stable core identity. A Nostalgia frame thus promises that returning to past values or conditions can restore comfort, meaning, or greatness.

Nostalgia also connects with the concept of collective memory~\cite{halbwachs1997memoire} where societies construct shared narratives of a golden age. These narratives foster social cohesion and increase optimism and resilience during periods of uncertainty. Empirical work suggests that nostalgia increases feelings of social connectedness and meaningfulness, functioning as an existential resource~\cite{wildschut2006nostalgia}. In political communication, nostalgic rhetoric is especially prominent in populist narratives because it is emotionally potent and identity-affirming~\cite{wodak2015politics,taggart2000populism}. By painting the past in positive hues, the Nostalgia frame activates longing for what is perceived as lost and motivates efforts to reclaim it.

\subsection{Temporal Contrast -- Then vs. Now and Temporal Comparison Theory}

The \emph{Temporal Contrast} frame juxtaposes past and present, highlighting change over time. This approach relates to the Temporal Comparison Theory, which extends social comparison principles to comparisons across time~\cite{albert1977temporal, suls2013handbook}. Just as individuals evaluate themselves relative to others, they also evaluate present circumstances relative to past or anticipated future states. Such comparisons make changes more salient by framing them relative to a reference point. This can amplify perceptions of progress, decline, or instability depending on the direction of contrast.

Temporal comparisons can be upward or downward. Upward temporal comparisons frame the past as better than the present, which elicits dissatisfaction, pessimism, or alarm. Downward temporal comparisons frame the past as worse, which promotes feelings of progress, relief, or hope~\cite{wills1981downward}. Statements such as ``once a thriving hub, now a city in decline'' exemplify upward temporal comparison.
These effects parallel findings in social comparison research, where upward comparisons can motivate or distress, and downward comparisons tend to bolster self-evaluation~\cite{wilson2001chump}. By placing two time points side by side, they invite comparative inferences about trajectory, decline, or progress. In this sense, Temporal Contrast frames draw persuasive force from a basic cognitive tendency toward reference-dependent evaluation where judgments are anchored to comparison points rather than being assessed in isolation \cite{prospect_theory}.

\subsection{Continuity -- Reassurance and Fatigue in Persistent Systems}

The \emph{Continuity} frame emphasizes persistence across time, encompassing stability, stagnation, momentum, or recurrence. It is realized by presenting conditions, processes, or trajectories as enduring, ongoing, or repeatedly recurring (e.g., ``X has been true for decades,'' ``history is repeating itself,'' or ``this trend continues''). While continuity can function to reassure or legitimize through longevity, we observed that in our corpus it is most frequently used to highlight stagnation, failure to progress, or the persistence of undesirable states.

From a cognitive and decision-science perspective, continuity connects to status quo bias, whereby individuals systematically prefer existing arrangements and resist change, even when alternatives may be superior \cite{samuelson1988status}. This bias, reinforced by loss aversion \cite{prospect_theory} and cognitive inertia \cite{mcguire1960cognitive}, helps explain why persistent conditions are often normalized and why continuity can be framed as inevitable rather than contestable. System justification theory further explains how endurance itself becomes a source of legitimacy, leading individuals to defend and rationalize long-standing social or institutional arrangements \cite{jost1994role}. Together, these mechanisms reinforce perceptions of stability, making persistent conditions appear natural, inevitable, or difficult to change meaningfully.

Beyond stagnation, continuity also manifests through cyclical or recurrent temporal models, in which repetition is framed as expected. When communicators invoke recurrence or cycles, continuity framing normalizes repetition and downplays the possibility of rupture or transformation. In political contexts, such framing can suggest that outcomes are unsurprising as they have already happened before, and are therefore historically patterned and hence are expected to continue to happen. Continuity can also express momentum. In this case, continuity framing highlights trends that continue to rise or decline by embedding them within an unfolding time trajectory. Such framing implies that change is either already underway or constrained by prior momentum.

Overall, Continuity framing derives persuasive force by presenting time as unbroken, patterned, or accumulating. Whether framed as legitimate endurance, inevitable recurrence, forward momentum, or frustrating stasis, continuity invites audiences to interpret the present as tightly constrained by what has persisted before.

\subsection{Skeptical -- Doubting the Future, Risk Aversion, and Negativity Bias}

The Skeptical frame casts doubt about the future (e.g., ``XYZ plan may collapse under pressure''), and its impact is correlated to several psychological tendencies related to risk perception and negative information processing. One key principle at play is loss aversion, a cornerstone of prospect theory~\cite{prospect_theory}. Loss aversion holds that people feel potential losses more acutely than equivalent gains. Therefore, a message focusing on what could go wrong or what one might lose if a scenario fails will often resonate strongly. By framing the future in terms of possible failure or pitfalls, the Skeptical frame leverages our predisposition to avoid loss and pain. Audiences are likely to give more weight to warnings of collapse than to equally plausible assurances of success because, psychologically, avoiding a bad outcome is more motivating than securing a good one. Closely related is the broader negativity bias in human cognition: negative information (warnings, dangers, criticisms) generally draws greater attention and has a bigger impact on judgments than positive information~\cite{kanouse1987negativity}. In communications, this is reflected in the news media's bias by injecting doubt and highlighting uncertainty about the future. 

Essentially, Skeptical framing tells the audience: do not get too comfortable, problems likely lie ahead. This can prompt more critical evaluation and vigilance. Another theoretical link is ambiguity aversion where people are uneasy about unknown outcomes and often prefer known risks over unknown ones~\cite{fox1995ambiguity}. A Skeptical frame accentuates the unknown (``will the plan collapse? It's uncertain''), potentially triggering a cautionary stance where people would rather err on the side of caution than bet on an overly optimistic plan. The framing derives its persuasive form by framing future developments as tenuous or doubtful, thus allowing communicators to justify conservative or preventive actions in the present.

\section{Additional Experimental Details}
\subsection{XLM-R Training Objective and Optimization Details}
\label{sec:training_details}

We next describe the optimization and training strategies used to address extreme label imbalance and sparsity in sentence-level temporal framing. In particular, we detail (i) a weighted Binary Cross-Entropy objective that corrects for frame-level imbalance across the corpus (Section~\ref{sec:wbce}), and (ii) a label-aware batching scheme that mitigates batch-level sparsity during training (Section~\ref{sec:lab}). While loss reweighting adjusts the relative contribution of rare temporal frames, label-aware batching ensures consistent exposure to positive supervision at the mini-batch level. Together, these mechanisms stabilize training in a multi-label setting where positive instances are both rare and unevenly distributed across frames.

\paragraph{Weighted Binary Cross-Entropy Objective}
\label{sec:wbce}

Let $K$ denote the number of temporal frames. For each sentence, the model produces logits $\mathbf{z} \in \mathbb{R}^K$, which are converted to probabilities via independent sigmoid activations. Ground-truth labels are represented as $\mathbf{y} \in \{0,1\}^K$, allowing multiple frames to be active simultaneously.

We optimize a weighted Binary Cross-Entropy loss defined as
\[
\mathcal{L}
= - \sum_{k=1}^{K}
\bigl[
w_k y_k \log \sigma(z_k)
+ (1 - y_k)\log(1 - \sigma(z_k))
\bigr]
\]

where $\sigma(\cdot)$ denotes the sigmoid function and $w_k$ is a positive class weight for frame $k$, computed from training label frequencies.

The weight $w_k$ is computed from the training data based on label prevalence:
\[
w_k = \frac{N - n_k}{n_k},
\]
where $N$ is the total number of training instances and $n_k$ is the number of sentences annotated with frame $k$. This weighting increases the contribution of rare temporal frames while leaving negative instances unweighted. The objective corresponds to standard Binary Cross-Entropy when $w_k = 1$.

\paragraph{Label-Aware Batching under Sparse Supervision}
\label{sec:lab}

In addition to loss reweighting, training is affected by extreme label sparsity: under naive random batching, many mini-batches contain no positive labels, resulting in weak or uninformative gradient signals. To address this, we employ a label-aware batching strategy that enforces a fixed proportion of temporally framed sentences per batch.

Concretely, each mini-batch is constructed to contain a predefined fraction of sentences with at least one active temporal frame, with remaining batch slots filled by non-framed sentences. This strategy ensures consistent exposure to positive supervision while preserving the overall distribution of negative instances. Importantly, this batching procedure affects only training and does not alter the evaluation distribution. In our experiments, we fix the proportion of temporally framed sentences to 50\% per mini-batch. Once all available positive instances have been exhausted, no additional batches are constructed for that epoch. Negative instances are sampled randomly without replacement. As a result, each training epoch may not cover the full dataset, and the set of negative sentences observed can vary across epochs, introducing additional stochasticity while maintaining balanced supervision across batches.

\paragraph{Hyperparameters and Randomness Control}
\label{sec:hyperparams}

All experiments uses the default set of hyperparameters unless stated otherwise. Models are fine-tuned using the AdamW optimizer with a learning rate of $2 \times 10^{-5}$ and a batch size of 16. Inputs are truncated to a maximum length of 512 tokens. Training proceeds for up to 20 epochs with early stopping based on development-set micro-averaged F1 score, using a patience of three epochs. The optimizer is instantiated with default AdamW parameters: $\beta_1 = 0.9$, $\beta_2 = 0.999$, $\epsilon = 10^{-8}$, and no weight decay applied.

To address label sparsity during training, we employ a label-aware batching strategy with a fixed batch size of 16, constructed to include 50\% sentences with at least one active temporal frame and 50\% non-framed sentences. Positive and negative instances are sampled without replacement, and batch construction stops once all positive instances have been exhausted for a given epoch. As a result, individual epochs may not cover the entire training set, and the sampled negative instances may vary across epochs slightly over time.

For reproducibility, we fix the random seed to 42 for all experiments. The seed is applied consistently across Python’s \texttt{random} module, NumPy, and PyTorch. All reported results correspond to the model checkpoint achieving the best performance on the development set under this fixed hyperparameter and seed configuration.

\subsection{LLM Experimental Details}

\subsubsection{Context Ablation}
\label{appx:llm:context_ablation}

Based on initial explorations with larger context windows using GPT-5.2 on the development set, we feed models only the target sentence itself as input, denoted \textsc{SENT} (sentence-only), in all experiments. This setup isolates sentence-level cues and avoids introducing noise from surrounding context. It also reduces token usage and inference cost. In this ablation, we compared \textsc{SENT} against two context-augmented variants: adding neighboring sentences (\textsc{NEIGH}) and providing the full article text (\textsc{DOC}). Adding context yielded no clear gains in overall detection performance and in some cases even reduced precision, likely due to irrelevant information. As illustrated in Table~\ref{tab:context_ablation}, we observed that surrounding context can make the model more conservative, leading it to miss frames that are already evident from the sentence alone. The sentence-only context achieves the highest performance while also being the most cost-effective. We, therefore, opt to use the \textsc{SENT} configuration for both fine-tuning and zero-shot evaluations.

\begin{table}[h]
\centering
\small
\resizebox{\columnwidth}{!}{%
\begin{tabular}{lccccc}
\toprule
Context Mode & Micro P & Micro R & Micro F1 & Macro F1 & Cost (\$) \\
\midrule
Sentence-only & 0.212 & 0.476 & \textbf{0.294} & 0.251 & 3.50 \\
Prev + Target + Next & 0.226 & 0.382 & 0.284 & 0.251 & 3.84 \\
Full document & 0.216 & 0.411 & 0.283 & 0.252 & 11.41 \\
\bottomrule
\end{tabular}%
}
\caption{\textbf{Effect of context granularity on zero-shot temporal framing classification.} Sentence-only input achieves the best performance while incurring the lowest cost, yielding the most efficient overall configuration.}
\label{tab:context_ablation}
\end{table}

\subsubsection{LLM Inference Details}
\label{appx:sec:llm_inference_details}

We use vLLM \cite{kwon2023EfficientMemoryManagement} for efficient batch inference of all LLM-based models, including both zero-shot and fine-tuned variants. vLLM enables high-throughput generation through continuous batching and PagedAttention. The random seed is fixed at 239 for reproducibility. The instruction set is shown in Appendix~\ref{sec:appendix_zeroshot}.

\paragraph{Sampling Configuration.}
We use consistent sampling parameters across all experiments: temperature $\tau=0.0$, top-p (nucleus sampling) $p=0.8$, and top-k $k=20$. The maximum generation length is set to 1024 tokens. Repetition, presence, and frequency penalties are disabled to avoid interfering with structured output generation.

\paragraph{Guided JSON Decoding.}
To ensure well-formed outputs, we employ vLLM's structured output feature with a JSON schema derived from a Pydantic model. The schema constrains the model to output a JSON object with a single key \texttt{temporal\_frames} containing a list of valid frame types. Valid frames are restricted to the eight categories: \emph{Continuity}, \emph{Nostalgia}, \emph{Primacy}, \emph{Recency}, \emph{Skeptical}, \emph{Temporal Anchoring}, \emph{Temporal Contrast}, and \emph{Urgency}. This guided decoding eliminates parsing errors from malformed JSON and prevents hallucinations.

\subsubsection{LLM Fine-Tuning Details}
\label{appx:sec:llm_finetuning_details}

We fine-tune Qwen3-8B and Llama-3.1-8B-Instruct using QLoRA~\citep{dettmers2023QLoRAEfficientFinetuning}, a parameter-efficient method that combines 4-bit quantization with low-rank adapters. Our implementation uses Unsloth\footnote{\url{https://github.com/unslothai/unsloth}}.

\paragraph{LoRA Configuration.}
We apply low-rank adapters to all attention and feed-forward modules: \texttt{q\_proj}, \texttt{k\_proj}, \texttt{v\_proj}, \texttt{o\_proj}, \texttt{gate\_proj}, \texttt{up\_proj}, and \texttt{down\_proj}. The LoRA hyperparameters are: rank $r=16$, scaling factor $\alpha=16$, and dropout $0.0$. We use Unsloth's gradient checkpointing for memory efficiency.

\paragraph{Training Configuration.}
Models are trained for 3 epochs with a per-device batch size of 16 and gradient accumulation over 4 steps, yielding an effective batch size of 64. We use the AdamW optimizer, a learning rate of $2 \times 10^{-4}$ with linear decay, and 10 warmup steps. Weight decay is set to 0.01 and gradient clipping to 1.0. The random seed is fixed at 239 for reproducibility.

\paragraph{Response-Only Training.}
Following best practices for instruction tuning, we apply response-only training, which masks the loss on instruction tokens and computes gradients only on the assistant's response during training.

\paragraph{Post-Training.}
After training, we merge the LoRA adapters into the base model weights using 16-bit precision (\texttt{merged\_16bit}), producing a standalone model compatible with vLLM inference. The merged models are used directly with our zero-shot inference pipeline to ensure consistent evaluation across all settings.

\subsection{Detailed Experimental Results}
\label{appx:sec:llm_experiment_details}

Following the results in Section~\ref{sec:results}, we provide additional experimental breakdowns by language and by temporal frame.
Table~\ref{tab:per_label_f1_all} reports per-label F1 scores for all zero-shot and supervised models, broken down by evaluation split (English+German, English, and German).
Table~\ref{tab:overall_results_by_language} shows binary detection and multilabel classification metrics under the same splits. The per-label breakdown reveals substantial variation across temporal frames.
Frames such as \emph{Continuity}, \emph{Temporal Anchoring}, and \emph{Temporal Contrast} are detected more reliably across models and languages, particularly in English, where these frames occur more frequently and are associated with relatively stable lexical or structural cues.
In contrast, low-frequency frames, most notably \emph{Nostalgia}, exhibit consistently low F1 scores across both zero-shot and supervised settings, which highlights the strong effect of data sparsity and the difficulty of learning frames that rely on implicit evaluative references to the past.

Comparing results across languages highlights the impact of corpus composition. Across models, English consistently outperforms German, especially for recall and rare frames, where the German subset contains far fewer positive instances. Consequently, lower German performance reflects data sparsity rather than a failure of cross-lingual modeling. Finally, the contrast between zero-shot and supervised models shows that task-specific training yields more balanced and consistent performance, though substantial per-frame variability remains, reflecting the discourse-level nature of temporal framing. These gains are most pronounced for higher-frequency frames, where supervision provides sufficient signal. However, improvements remain limited for rare and implicit frames, underscoring the persistent challenge of data sparsity.

\onecolumn
\newpage
\noindent
\begin{minipage}{\linewidth}
\centering

{\fontsize{7.7}{8.7}\selectfont
\setlength{\tabcolsep}{10pt}
\renewcommand{\arraystretch}{0.9}
\scalebox{0.97}{%
\begin{tabular}{llcccccccc}
\toprule
\textbf{Split} & \textbf{Model} &
Pri & Rec & Urg & Anch & Nos & Ctr & Cnt & Skp \\
\midrule

\multirow{9}{*}{EN + DE}
& \textit{Zero-shot} \\
& LLaMA-3.1-8B   & 0.11 & 0.20 & 0.06 & 0.13 & 0.10 & 0.14 & 0.16 & 0.00 \\
& LLaMA-3.3-70B  & 0.16 & 0.17 & 0.24 & 0.21 & 0.19 & 0.23 & 0.33 & 0.17 \\
& Qwen3-8B       & 0.03 & 0.10 & 0.16 & 0.15 & 0.11 & 0.23 & 0.27 & 0.07 \\
& Qwen3-32B      & 0.19 & 0.28 & 0.18 & 0.26 & 0.16 & 0.23 & 0.31 & 0.18 \\
& Qwen3-235B     & 0.16 & 0.22 & 0.21 & 0.24 & 0.32 & 0.26 & 0.30 & 0.12 \\
& GPT-5.2        & 0.18 & 0.25 & 0.20 & 0.23 & 0.07 & 0.27 & 0.29 & 0.11 \\
& \textit{Supervised} \\
& LLaMA-3.1-8B-Instruct & 0.41 & 0.45 & 0.32 & 0.38 & 0.00 & 0.42 & 0.50 & 0.34 \\
& Qwen3-8B              & 0.39 & 0.49 & 0.31 & 0.34 & 0.00 & 0.41 & 0.55 & 0.28 \\

\midrule
\multirow{9}{*}{EN}
& \textit{Zero-shot} \\
& LLaMA-3.1-8B   & 0.16 & 0.28 & 0.12 & 0.25 & 0.19 & 0.20 & 0.22 & 0.00 \\
& LLaMA-3.3-70B  & 0.20 & 0.25 & 0.30 & 0.28 & 0.30 & 0.42 & 0.40 & 0.17 \\
& Qwen3-8B       & 0.08 & 0.21 & 0.25 & 0.25 & 0.26 & 0.30 & 0.35 & 0.05 \\
& Qwen3-32B      & 0.28 & 0.44 & 0.25 & 0.34 & 0.19 & 0.29 & 0.32 & 0.10 \\
& Qwen3-235B     & 0.26 & 0.36 & 0.29 & 0.37 & 0.38 & 0.36 & 0.36 & 0.05 \\
& GPT-5.2        & 0.23 & 0.31 & 0.28 & 0.33 & 0.18 & 0.36 & 0.41 & 0.15 \\
& \textit{Supervised} \\
& LLaMA-3.1-8B-Instruct & 0.34 & 0.49 & 0.34 & 0.42 & 0.00 & 0.46 & 0.53 & 0.34 \\
& Qwen3-8B              & 0.43 & 0.52 & 0.24 & 0.40 & 0.00 & 0.48 & 0.60 & 0.22 \\

\midrule
\multirow{9}{*}{DE}
& \textit{Zero-shot} \\
& LLaMA-3.1-8B   & 0.06 & 0.06 & 0.02 & 0.04 & 0.06 & 0.00 & 0.05 & 0.00 \\
& LLaMA-3.3-70B  & 0.09 & 0.06 & 0.14 & 0.14 & 0.13 & 0.08 & 0.20 & 0.18 \\
& Qwen3-8B       & 0.01 & 0.03 & 0.08 & 0.07 & 0.04 & 0.13 & 0.16 & 0.08 \\
& Qwen3-32B      & 0.10 & 0.09 & 0.09 & 0.17 & 0.12 & 0.10 & 0.28 & 0.29 \\
& Qwen3-235B     & 0.06 & 0.07 & 0.13 & 0.13 & 0.27 & 0.16 & 0.18 & 0.23 \\
& GPT-5.2        & 0.11 & 0.14 & 0.12 & 0.13 & 0.00 & 0.16 & 0.17 & 0.08 \\
& \textit{Supervised} \\
& LLaMA-3.1-8B-Instruct & 0.67 & 0.20 & 0.27 & 0.10 & 0.00 & 0.20 & 0.33 & 0.35 \\
& Qwen3-8B              & 0.30 & 0.36 & 0.48 & 0.07 & 0.00 & 0.26 & 0.40 & 0.40 \\
\bottomrule
\end{tabular}
}
}

\vspace{4pt}
\captionof{table}{\textbf{Per-label F1 scores across language evaluation splits.} Cnt = \emph{Continuity}, Nos = \emph{Nostalgia}, Pri = \emph{Primacy}, Rec = \emph{Recency}, Skp = \emph{Skeptical}, Anch = \emph{Temporal Anchoring}, Ctr = \emph{Temporal Contrast}, Urg = \emph{Urgency}.}
\label{tab:per_label_f1_all}

\vspace{20pt}

{\fontsize{7.7}{8.7}\selectfont
\setlength{\tabcolsep}{12pt}
\renewcommand{\arraystretch}{0.9}
\scalebox{0.97}{%
\begin{tabular}{llccc|cccc}
\toprule
\textbf{Split} & \textbf{Model} &
\multicolumn{3}{c}{\textbf{Binary Detection}} &
\multicolumn{4}{c}{\textbf{Multilabel Classification}} \\
\cmidrule(lr){3-5} \cmidrule(lr){6-9}
& & P & R & F1 & P & R & Micro-F1 & Macro-F1 \\
\midrule
\multirow{9}{*}{EN + DE}
& Random Baseline & 0.11 & 0.88 & 0.20 & 0.02 & 0.49 & 0.04 & 0.03 \\

\cmidrule(lr){2-9}
& \multicolumn{8}{l}{\textit{Zero-shot models}} \\
& LLaMA-3.1-8B & 0.14 & 0.94 & 0.24 & 0.06 & 0.34 & 0.10 & 0.11 \\
& LLaMA-3.3-70B & 0.28 & 0.78 & 0.41 & 0.16 & 0.43 & 0.23 & 0.21 \\
& Qwen3-8B & 0.21 & 0.75 & 0.33 & 0.08 & 0.29 & 0.13 & 0.14 \\
& Qwen3-32B & 0.36 & 0.57 & 0.44 & 0.20 & 0.31 & 0.25 & 0.22 \\
& Qwen3-235B & 0.31 & 0.70 & 0.44 & 0.17 & 0.38 & 0.24 & 0.23 \\
& GPT-5.2 & 0.34 & 0.67 & 0.45 & 0.24 & 0.46 & 0.31 & 0.29 \\

\cmidrule(lr){2-9}
& \multicolumn{8}{l}{\textit{Supervised fine-tuning}} \\
& XLM-R & 0.38 & 0.75 & 0.51 & 0.28 & 0.57 & 0.37 & 0.33 \\
& Llama-3.1-8B-Instruct & 0.67 & 0.45 & 0.54 & 0.48 & 0.37 & 0.42 & 0.35 \\
& Qwen3-8B & 0.60 & 0.54 & 0.57 & 0.48 & 0.40 & 0.44 & 0.35 \\

\midrule
\multirow{9}{*}{EN}
& Random Baseline & 0.20 & 0.88 & 0.32 & 0.03 & 0.48 & 0.06 & 0.06 \\

\cmidrule(lr){2-9}
& \multicolumn{8}{l}{\textit{Zero-shot models}} \\
& LLaMA-3.1-8B & 0.25 & 0.92 & 0.39 & 0.12 & 0.37 & 0.19 & 0.18 \\
& LLaMA-3.3-70B & 0.43 & 0.73 & 0.54 & 0.26 & 0.43 & 0.32 & 0.29 \\
& Qwen3-8B & 0.41 & 0.69 & 0.52 & 0.18 & 0.29 & 0.22 & 0.22 \\
& Qwen3-32B & 0.49 & 0.51 & 0.50 & 0.32 & 0.29 & 0.31 & 0.28 \\
& Qwen3-235B & 0.49 & 0.64 & 0.56 & 0.30 & 0.36 & 0.33 & 0.30 \\
& GPT-5.2 & 0.46 & 0.58 & 0.52 & 0.37 & 0.44 & 0.40 & 0.26 \\

\cmidrule(lr){2-9}
& \multicolumn{8}{l}{\textit{Supervised fine-tuning}} \\
& LLaMA-3.1-8B & 0.68 & 0.52 & 0.59 & 0.48 & 0.42 & 0.45 & 0.36 \\
& Qwen3-8B & 0.67 & 0.58 & 0.62 & 0.54 & 0.42 & 0.47 & 0.36 \\

\midrule
\multirow{9}{*}{DE}
& Random Baseline & 0.05 & 0.90 & 0.10 & 0.01 & 0.53 & 0.02 & 0.02 \\

\cmidrule(lr){2-9}
& \multicolumn{8}{l}{\textit{Zero-shot models}} \\
& LLaMA-3.1-8B & 0.07 & 0.99 & 0.12 & 0.02 & 0.25 & 0.03 & 0.04 \\
& LLaMA-3.3-70B & 0.16 & 0.91 & 0.27 & 0.07 & 0.45 & 0.12 & 0.13 \\
& Qwen3-8B & 0.11 & 0.93 & 0.19 & 0.03 & 0.30 & 0.05 & 0.07 \\
& Qwen3-32B & 0.23 & 0.72 & 0.35 & 0.10 & 0.37 & 0.16 & 0.15 \\
& Qwen3-235B & 0.18 & 0.85 & 0.29 & 0.08 & 0.43 & 0.13 & 0.15 \\
& GPT-5.2 & 0.21 & 0.83 & 0.33 & 0.14 & 0.64 & 0.23 & 0.17 \\

\cmidrule(lr){2-9}
& \multicolumn{8}{l}{\textit{Supervised fine-tuning}} \\
& LLaMA-3.1-8B & 0.61 & 0.24 & 0.34 & 0.46 & 0.20 & 0.28 & 0.26 \\
& Qwen3-8B & 0.43 & 0.43 & 0.43 & 0.33 & 0.33 & 0.33 & 0.28 \\
\bottomrule
\end{tabular}
}
}
\vspace{4pt}
\captionof{table}{\textbf{Overall results broken down by language evaluation splits.}}
\label{tab:overall_results_by_language}

\end{minipage}
\clearpage

\section{Full Annotated Example}
\label{sec:full_annotated_example}



\noindent\makebox[\linewidth]{%
  \fbox{%
    \parbox{0.95\textwidth}{%
      \scriptsize
      \setlength{\parindent}{0pt}
      \setlength{\parskip}{0.35em}
      \setlength{\baselineskip}{0.92\baselineskip}

      {\centering \textbf{Violence Has Been Normalized}\par}

During the misnamed and mostly preposterous debate between Kamala Harris and Donald Trump, a moderator fact-checked Trump’s claim that crime is up.

In contrast to Trump’s claim, moderator David Muir said that the FBI reports that crime is down, a claim that likely struck every viewer as obviously wrong.

Shoplifting was not a way of life before lockdowns.\tfsep
\tflabel{Nostalgia}{tfNostalgia}\tfsep
\tflabel{Temporal Anchoring}{tfAnchoring}\tfsep
\tflabel{Temporal Contrast}{tfContrast}\tfsentgap
Most cities were not demographic minefields of danger around every corner.\tfsep
\tflabel{Nostalgia}{tfNostalgia}\tfsep
\tflabel{Temporal Contrast}{tfContrast}\tfsentgap
There was no such thing as a drugstore with nearly all products behind locked Plexiglas.\tfsep
\tflabel{Nostalgia}{tfNostalgia}\tfsep
\tflabel{Temporal Contrast}{tfContrast}

We weren’t warned of spots in cities, even medium-sized ones, where carjacking was a real risk.\tfsep
\tflabel{Nostalgia}{tfNostalgia}\tfsep
\tflabel{Temporal Contrast}{tfContrast}

It is wildly obvious that high crime in the U.S. is endemic, with ever less respect for person and property.\tfsep
\tflabel{Continuity}{tfContinuity}\tfsep
\tflabel{Primacy}{tfPrimacy}\tfsentgap
As for the FBI’s statistics, they’re worth about as much as most data coming from federal agencies these days.\tfsep
\tflabel{Temporal Contrast}{tfContrast}

They’re there for purposes of propaganda, manipulated to present the most favorable picture possible to help the regime.
Lies, Damn Lies and Government Statistics.

This is certainly true of the Bureau of Labor Statistics and the Commerce Department, which have been shoveling out obvious nonsense for years.\tfsep
\tflabel{Continuity}{tfContinuity}\tfsentgap
Professionals in the field know it but go along for reasons of professional survival.\tfsentgap
In truth, we’ve never had a real economic recovery since lockdowns.\tfsep
\tflabel{Continuity}{tfContinuity}\tfsep
\tflabel{Temporal Anchoring}{tfAnchoring}

Crime is up.\tfsep
\tflabel{Temporal Contrast}{tfContrast}\tfsentgap
Literacy is down.\tfsep
\tflabel{Temporal Contrast}{tfContrast}\tfsentgap
Trust has collapsed.\tfsep
\tflabel{Temporal Contrast}{tfContrast}\tfsentgap
Societies were shattered and remain so.\tfsep
\tflabel{Continuity}{tfContinuity}\tfsep
\tflabel{Temporal Contrast}{tfContrast}

Only a few weeks following the officious fact-check at the debate, we now have new data from the National Crime Victimization Survey.\tfsep
\tflabel{Recency}{tfRecency}\tfsentgap
The Wall Street Journal reports: “The urban violent-crime rate increased 40\% from 2019--2023.
Excluding simple assault, the urban violent-crime rate rose 54\% over that span.
From 2022--2023, the urban violent-crime rate didn’t change to a statistically significant degree, so these higher crime rates appear to be the new norm in America’s cities.”

But the FBI tries to tell you that crime is down.
Sure, whatever they say.

The report isolates the “post-George Floyd protests” because no media source wants to mention the lockdowns.\tfsep
\tflabel{Temporal Anchoring}{tfAnchoring}\tfsentgap
It is still a taboo subject.\tfsep
\tflabel{Continuity}{tfContinuity}

We somehow cannot say, even now, that the worst abuses of rights in U.S. history in terms of scale and depth were a disaster, simply because saying so implicates the whole of the media, both parties, all government agencies, academia and all the upper reaches of the social and political order.

\textbf{Politics Has Become Life and Death}

The problem of political division is getting alarmingly serious.\tfsep
\tflabel{Continuity}{tfContinuity}\tfsep
\tflabel{Urgency}{tfUrgency}\tfsentgap
It’s no longer just about competing yard signs and loud rallies.\tfsep
\tflabel{Temporal Contrast}{tfContrast}\tfsentgap
We now have regular assassination attempts, plus even an extremely strange appearance of a bounty put on a candidate’s head by an official agency.\tfsep
\tflabel{Temporal Contrast}{tfContrast}

Surveys have shown that 26 million people in the U.S. believe that violence is fine to keep Trump from regaining the presidency.\tfsep
\tflabel{Continuity}{tfContinuity}\tfsentgap
Where might people have gotten that idea?

Probably from many Hollywood movies that fantasize about having killed Hitler before he accomplished his evil plus the nonstop likening of Trump to Hitler, and hence one follows from another.
Liken Trump to Hitler and that is the result you produce.

There’s private violence, public violence and many forms in between including vigilante violence.
Rights violations against person and property are now normalized.\tfsep
\tflabel{Temporal Contrast}{tfContrast}\tfsentgap
This springs from the culture of our times which has been heavily informed and even defined by the deployment of state violence in service of policy goals, at a scale, scope and depth never before seen.\tfsep
\tflabel{Continuity}{tfContinuity}\tfsep
\tflabel{Primacy}{tfPrimacy}\tfsep
\tflabel{Temporal Contrast}{tfContrast}

\textbf{The Role of Censorship}

Censorship is a major part of it.
Censorship is the deployment of force in service of state power, and other institutions connected to state power, for purposes of culture planning.
It’s exercised by the shallow state, in response to the middle state, and on behalf of the deep state.
It’s a form of violence that interrupts the free flow of information: the ability to speak, and the ability to learn.
Censorship trains the population to be quiet, afraid and constantly stressed, and it sorts people by the compliant versus the dissidents.
Censorship is designed to shape the public mind toward the end of shoring up regime stability.
Once it starts, there’s no limit to it.\tfsep
\tflabel{Continuity}{tfContinuity}

I’ve mentioned to people that Substack, Rumble and X could be banned by the spring of next year, and people respond with incredulity.\tfsep
\tflabel{Skeptical}{tfSkeptical}\tfsentgap
Why?

Four years ago, we were locked in our homes and locked out of churches, and the schools for which people pay all year were shut down by government force.\tfsep
\tflabel{Temporal Anchoring}{tfAnchoring}\tfsentgap
If they can do that, they can do anything.

\textbf{Remember Free Speech?}\tfsep\tflabel{Nostalgia}{tfNostalgia}

Censorship has been so effective that it’s changed the way we engage with each other even in private.\tfsep
\tflabel{Continuity}{tfContinuity}\tfsep
\tflabel{Temporal Contrast}{tfContrast}

Brownstone Institute, which I founded, recently held a private retreat for scholars, fellows and special guests.
One very special guest wrote me that she was completely shocked at the freedom of thought and speech that was present in the room.
As a mover in the highest circles, she had forgotten what that was like.

This censorship coincides with a strange valorization of violence that we are presented with from all over the world: Ukraine, the Middle East, London, Paris and many American cities.
Never have so many held video cameras in their pockets and never have there been so many platforms on which to post the results.\tfsep
\tflabel{Primacy}{tfPrimacy}\tfsep
\tflabel{Temporal Contrast}{tfContrast}
One does wonder how all these relentless presentations of destruction and killing affect public culture.\tfsep
\tflabel{Continuity}{tfContinuity}

\textbf{Why They’re Doing It}

What purpose are all these soft, hard, public and private exercises of violence serving?
The standard of living is suffering, lives are shortening, despair and ill health are main features of the population and illiteracy has swept through an entire generation.\tfsep
\tflabel{Continuity}{tfContinuity}

The decision to deploy violence to master the microbial kingdom did not turn out well.
Worse, it unleashed violence as a way of life.

“When plunder becomes a way of life for a group of men in a society,” wrote Frederic Bastiat, “over the course of time they create for themselves a legal system that authorizes it and a moral code that glorifies it.”

That is precisely where we are.
It’s time we talk about it and name the culprit.\tfsep
\tflabel{Urgency}{tfUrgency}
Liberty, privacy and property were already unsafe before 2020 but it was the lockdowns that unleashed Pandora’s box of evils.\tfsep
\tflabel{Continuity}{tfContinuity}\tfsep
\tflabel{Temporal Anchoring}{tfAnchoring}\tfsep
\tflabel{Temporal Contrast}{tfContrast}

We cannot live this way.
The only arguments worth having are those that name the reason for the suffering and offer a viable path back to civilized living.\tfsep
\tflabel{Nostalgia}{tfNostalgia}

}%
}
}

\medskip
\captionof{figure}{\textbf{A full annotated example illustrating temporal framing.} Adapted from ``Violence Has Been Normalized'' by Jeffrey Tucker, originally published on \textit{Daily Reckoning} (Oct.~11,~2024). 
URL: \url{https://dailyreckoning.com/violence-has-been-normalized/}.}%

\clearpage
\section{Prompt for Zero-Shot Experiments}
\label{sec:appendix_zeroshot}

\vspace{-0.6em}

\begin{center}
  \begin{framed}
  \noindent
  \begin{minipage}{\linewidth}
  \begin{Verbatim}[fontsize=\footnotesize,  baselinestretch=0.88,
commandchars=\\\{\}]
\darkbluetext{Multi-label Temporal Framing Classification}

\lightbrowntext{You are an expert annotator for temporal framing in news text. Temporal framing refers to the}
\lightbrowntext{rhetorical use of time-related elements to persuade, not merely to report chronology. A}
\lightbrowntext{sentence should be labeled only if the temporal element contributes to its persuasive force.}

\lightbrowntext{Temporal framing taxonomy (short definitions and examples):}
\darkbluetext{\{}\lightbluetext{BRIEF\_TAXONOMY}\darkbluetext{\}}

\lightbrowntext{Key rules for annotation:}
\lightbrowntext{- Do not annotate factual chronological reporting or purely descriptive temporal mentions.}
\lightbrowntext{- Do not annotate quoted or indirect speech.}
\lightbrowntext{- If removing the temporal expression does not change the persuasive force, do not annotate.}
\lightbrowntext{- Multiple temporal frames may apply to a single sentence.}

\lightbrowntext{Context \contexttext{(optional)}:}
\darkbluetext{\{}\lightbluetext{CONTEXT}\darkbluetext{\}}

\lightbrowntext{Target sentence:}
\darkbluetext{\{}\lightbluetext{TARGET\_SENTENCE}\darkbluetext{\}}

\lightbrowntext{Return the applicable temporal frames in structured JSON.}

  \end{Verbatim}
  \end{minipage}
  \end{framed}
\captionof{figure}{
\textbf{Prompt template for temporal framing classification.}
Taxonomy definitions from Figure~\ref{tab:temporal-framing-definitions} are inserted into the prompt via the \texttt{BRIEF\_TAXONOMY} variable, after which the model predicts all applicable temporal frames in a single inference pass.
The optional context block is included only in context-granularity ablation experiments.
}
  \label{fig:single_step_prompt}
\end{center}

\section{INCEpTION Annotation Platform}
\label{sec:inception}

All annotations in this work were produced using the INCEpTION annotation platform, a web-based system designed for scalable, multi-layer linguistic annotation. INCEpTION supports span-based and sentence-level labeling, and configurable tagsets. These features make it suitable for complex annotation tasks that require consistency and quality control across multiple annotators and languages.

For temporal framing, we configured INCEpTION to operate at the sentence level, where annotators assign one or more temporal frame labels to a sentence. The interface allows annotators to view the full article context while focusing annotation decisions on individual sentences. The interface further facilitates workload assignment and monitoring, as well as inspection, revision, and adjudication of annotations.

\begin{figure}[th]
    \centering
    \includegraphics[width=\linewidth]{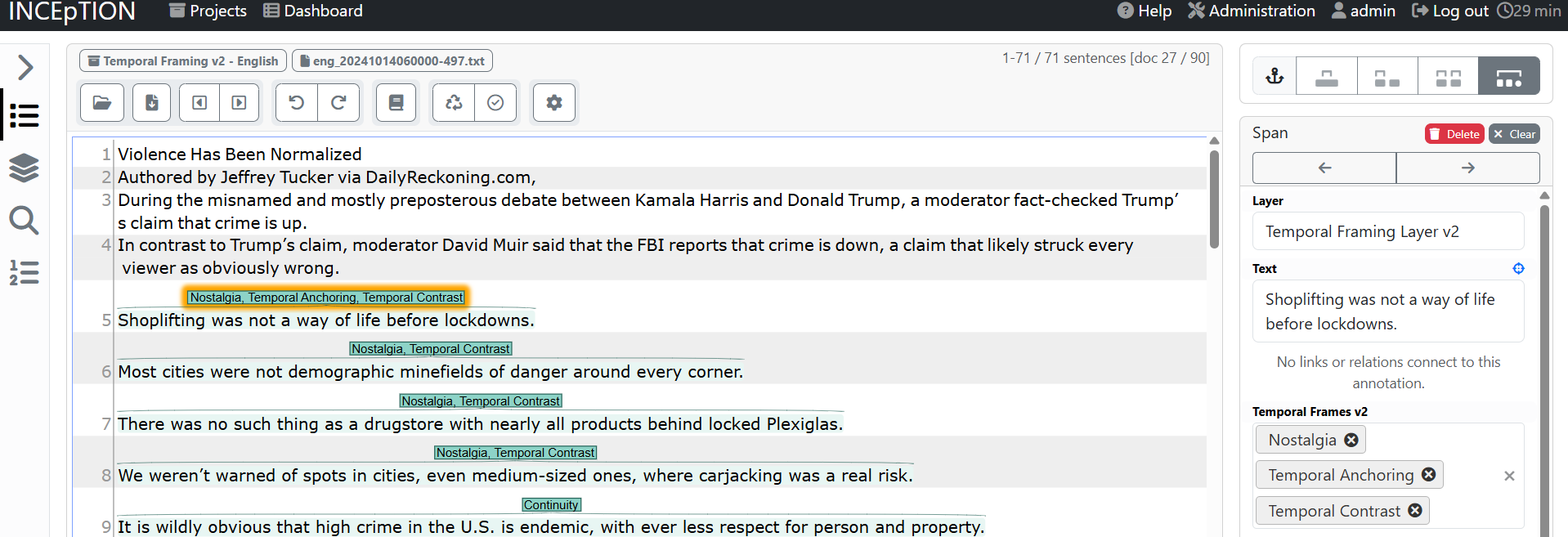}
    \caption{\textbf{INCEpTION annotation interface.}
    The figure shows how the annotated example shown in Figure~\ref{fig:temporal_framing_inline_example} looks like on the INCEpTION interface with sentence-level annotation and multiple temporal frame labels.}
    \label{fig:annotation_ui}
\end{figure}

Figure~\ref{fig:annotation_ui} shows the annotation interface as used in this work.

\end{document}